\documentclass{article}

\usepackage{microtype}
\usepackage{graphicx}
\usepackage{booktabs} % for professional tables

\usepackage{hyperref}

\usepackage[accepted]{icml2024}

\usepackage{amsmath}
\usepackage{amssymb}
\usepackage{mathtools}
\usepackage{amsthm}
\usepackage{booktabs} % commands to create good-looking tables
\usepackage{tikz} % nice language for creating drawings and diagrams
\usepackage{times}
\usepackage{epsfig}
\usepackage{graphicx}
\usepackage{amsmath}
\usepackage{amssymb}
\usepackage{booktabs}
\usepackage{multirow}
\usepackage{newfloat}
\usepackage{listings}
\usepackage{colortbl}
\usepackage{multicol}
\usepackage{enumitem}
\usepackage{color, colortbl}
\usepackage{comment}
\usepackage{nicefrac}
\usepackage{pifont}
\usepackage{caption}
\usepackage{stfloats}
\usepackage{subcaption}

\definecolor{Gray}{gray}{.9}
\usepackage{pgfplotstable}
\newcommand{\cmark}{\ding{51}}%
\newcommand{\xmark}{\ding{55}}%
\newcolumntype{g}{>{\columncolor{Gray}}c}
\usepackage[capitalize,noabbrev]{cleveref}

\theoremstyle{plain}

\theoremstyle{definition}

\theoremstyle{remark}

\usepackage[textsize=tiny]{todonotes}
\newcommand{\ie}{i.e.,}
\newcommand{\eg}{e.g.,}
\definecolor{GTgreen}{HTML}{459948}

\icmltitlerunning{Evaluation of Test-Time Adaptation Under Computational Time Constraints}

\begin{document}

\twocolumn[
\icmltitle{Evaluation of Test-Time Adaptation Under Computational Time Constraints}



\icmlsetsymbol{equal}{*}
% \and
% Hani Itani$^{1}$
% \and 
% Alejandro Pardo$^{1}$
% \and
% Shyma Alhuwaider$^{1}$
% \and
% Merey Ramazanova$^{1}$
% \and
% Juan C. Pérez$^{1}$
% \and
% Zhipeng Cai$^{2}$
% \and
% Matthias Müller$^{2}$
% \and 
% Bernard Ghanem$^{1}$\vspace{0.1cm}\and 
% $^1$ King Abdullah University of Science and Technology (KAUST). $^2$ Intel Labs.   
\begin{icmlauthorlist}
\icmlauthor{Motasem Alfarra}{kaust,intel}
\icmlauthor{Hani Itani}{kaust}
\icmlauthor{Alejandro Pardo}{kaust}
\icmlauthor{Shyma Alhuwaider}{kaust}
\icmlauthor{Merey Ramazanova}{kaust}
\icmlauthor{Juan C. Pérez}{kaust}
\icmlauthor{Zhipeng Cai}{intel}
\icmlauthor{Matthias Müller}{intel}
\icmlauthor{Bernard Ghanem}{kaust}
\end{icmlauthorlist}

\icmlaffiliation{kaust}{King Abdullah University of Science and Technology (KAUST), Thuwal, Saudi Arabia}
\icmlaffiliation{intel}{Intel Labs, Munich, Germany}

\icmlcorrespondingauthor{Motasem Alfarra}{motasem.alfarra@kaust.edu.sa}

\icmlkeywords{Machine Learning, ICML}

\vskip 0.3in
]



\printAffiliationsAndNotice{}  % leave blank if no need to mention equal contribution

\begin{abstract}

This paper proposes a novel online evaluation protocol for Test Time Adaptation (TTA) methods, which penalizes slower methods by providing them with fewer samples for adaptation. 
TTA methods leverage unlabeled data at test time to adapt to distribution shifts. 
Although many effective methods have been proposed, their impressive performance usually comes at the cost of significantly increased computation budgets. 
Current evaluation protocols overlook the effect of this extra computation cost, affecting their real-world applicability. 
To address this issue, we propose a more realistic evaluation protocol for TTA methods, where data is received in an online fashion from a constant-speed data stream, thereby accounting for the method's adaptation speed. 
We apply our proposed protocol to benchmark several TTA methods on multiple datasets and scenarios. 
Extensive experiments show that, when accounting for inference speed, simple and fast approaches can outperform more sophisticated but slower methods. 
For example, SHOT from 2020, outperforms the state-of-the-art method SAR from 2023 in this setting. 
Our results reveal the importance of developing practical TTA methods that are both accurate and efficient\footnote{Code: \href{https://github.com/MotasemAlfarra/Online_Test_Time_Adaptation}{github/MotasemAlfarra/Online-Test-Time-Adaptation}}.

% emphasizes the need for developing TTA methods that are efficient in realistic settings.
% \footnote{Code: \href{https://github.com/MotasemAlfarra/Online_Test_Time_Adaptation}{github.com/MotasemAlfarra/Online-Test-Time-Adaptation}\\  Corresponding to: \texttt{motasem.alfarra@kaust.edu.sa}}
%\footnote{Source code will be released upon acceptance.}.

% can we claim our protocol solidifies an incentive to make efficient mehtods?

\end{abstract}
\section{Introduction}
In recent years, Deep Neural Networks (DNNs) have demonstrated remarkable success in various tasks~\cite{he2016deep} thanks to their ability to learn from large datasets~\cite{deng2009imagenet}. 
% Deep neural networks~(DNNs) trained on large datasets have achieved significant success in various tasks~\cite{he2016deep}. 
However, a significant limitation of DNNs is their poor performance when tested on out-of-distribution data, which violates the i.i.d. assumption that the training and testing data are from the same distribution ~\cite{imagenetr,imagenetc,3dcc}. 
% However, DNNs fail when tested on out-of-distribution samples that violate the i.i.d. assumption~\cite{imagenetc,imagenetr}, whereby the training and testing distributions are identical. 
% Adversarial attacks, for example, are exploiting this distribution shift and have led to the emergence of a whole new field studying the robustness of DNNs ~\cite{x}. 
Such failure cases are concerning, since distribution shifts are common in real-world applications, \emph{e.g.}, image corruptions~\cite{imagenetc}, changing weather conditions~\cite{sakaridis2021acdc}, or security breaches~\cite{goodfellow2014explaining}. % among others~\cite{}. 
% Thus, applications leveraging DNNs in the real world require techniques to handle distribution shifts.
\begin{figure}
    \centering
    \includegraphics[width=0.95\columnwidth]{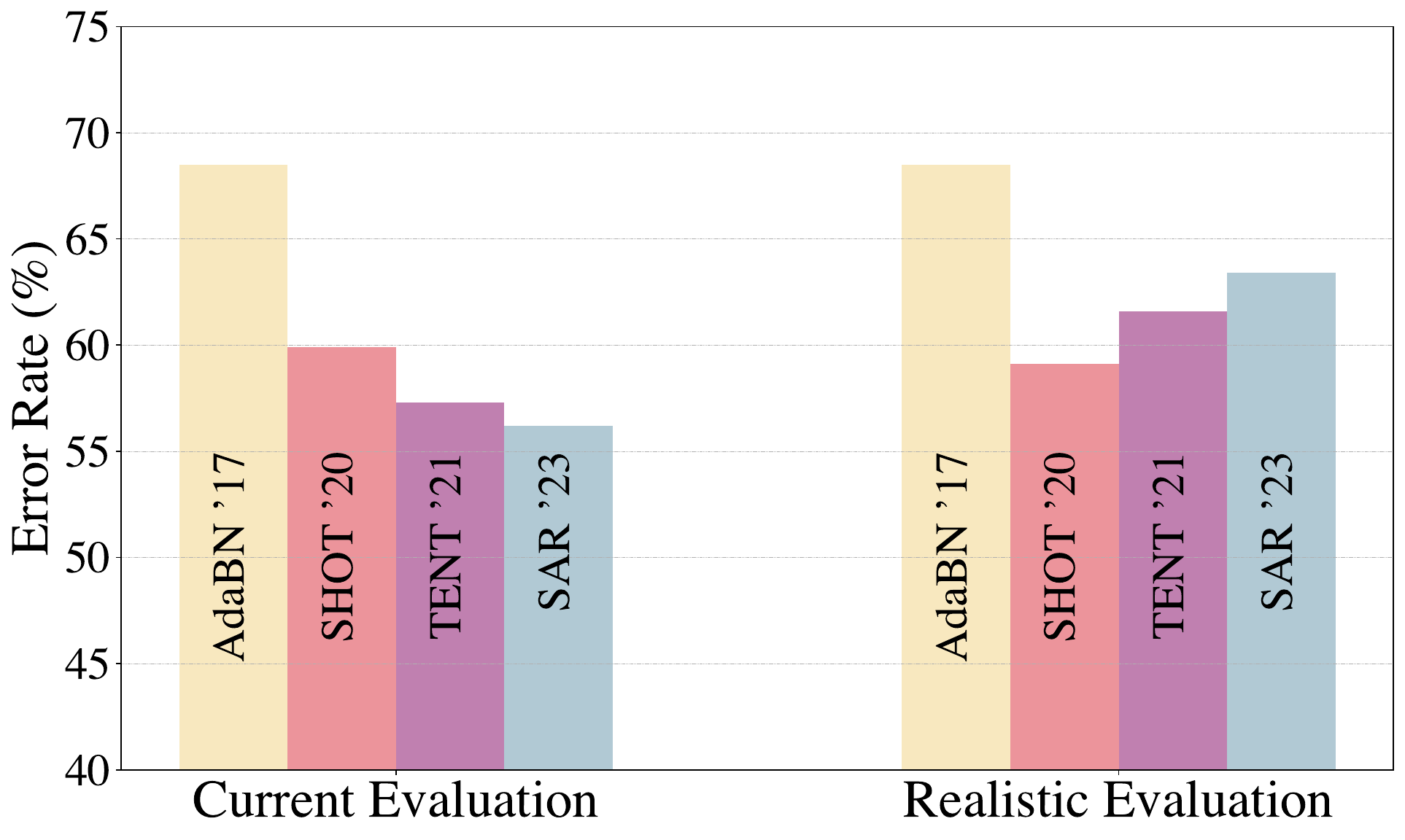}\vspace{-0.15cm}
    \caption{\textbf{The trend of average error rate using \textit{offline} evaluation \emph{vs} our proposed \textit{online} evaluation.} 
    In the offline setup, TTA methods demonstrate progress across time with a decreasing average error rate, \emph{e.g.} from $68.5\%$ using AdaBN to $56.2\%$ using SAR. 
    We propose a realistic evaluation protocol that accounts for the adaptation speed of TTA methods. 
    Under this protocol, fast methods (\emph{e.g.} AdaBN) are unaffected, while slower (but more recent and sophisticated) methods (\emph{e.g.} SAR) are penalized. 
    }
\label{fig:pull_fig}\vspace{-0.25cm}
\end{figure}

% \begin{figure}
%     \centering
%     \includegraphics[width=0.95\columnwidth]{Figures/PullFig/pullfig_final.pdf}\vspace{-0.15cm}
%     \caption{\textbf{The trend of average error rate using \textit{offline} evaluation \emph{vs} our proposed \textit{online} evaluation.} 
%     In the offline setup, TTA methods demonstrate progress across time with a decreasing average error rate, \emph{e.g.} from $68.5\%$ using AdaBN~\cite{adabn} to $56.2\%$ using SAR~\cite{sar}. 
%     We propose an online evaluation protocol that accounts for the adaptation speed of TTA methods. 
%     Under this protocol, fast methods (\emph{e.g.} AdaBN) are unaffected, while slower (but more recent and sophisticated) methods (\emph{e.g.} SAR) are penalized. 
%     }
%     \label{fig:pull_fig}\vspace{-0.25cm}
% \end{figure}
% \begin{figure}
%     \centering
%     \includegraphics[width=0.95\columnwidth,height=0.3\textwidth]{Figures/PullFig/pullfig_v4.pdf}
%     \caption{\textbf{Sample pull fig.} avg vals for eps on imgntc Bla bla bla.}
%     \label{fig:pull_fig}
% \end{figure}

Test Time Adaptation (TTA) ~\cite{saenko2010adapting, ttt, liu2021ttt++} has demonstrated promising results for solving the above problem.
TTA leverages the unlabeled data that arrives at test time by adapting the forward pass of pre-trained DNNs according to some proxy task~\cite{shot,pl}.
Though recent methods have made significant progress at improving accuracy under distribution shifts~\cite{tent,eata,dda}, many of them incur high computational overhead. 
For instance, some methods require self-supervised fine-tuning on the data~\cite{chen2022contrastive}, while others perform a diffusion process per input~\cite{dda}. % s before the DNN's forward pass~\cite{dda}. 
% \merey{if we have just 1 example of these methods, mb we can just put it there directly (DDA runs a diffusion...}
% Such extra computational cost can in practice decrease the inference speed, which is an important constraint in many applications. However, the current TTA methods use evaluation protocols that overlook this practical issue. 

The computational overhead of TTA methods decreases their inference speed, which is a critical property in many real-world applications that require the TTA method to produce predictions at the speed of the stream itself. 
This property, however, is overlooked in the current evaluation protocols for TTA methods. % overlooks this practical issue by evaluating them in an offline fashion.
In particular, these protocols assume a setting, which neglects how % the practical impact that decreased inference speed has in real-world scenarios. %  where domain shifts occur.
events constantly unfold regardless of the model's speed, causing the model to miss incoming samples when it is busy processing previous ones.
% In practice, events unfold and are presented to the model as a constant \textit{stream} of data, irrespective of the model's inference speed.
% This setup has practical costs as the model's processing speed determines the number of incoming samples that may be missed while it is busy processing previous samples.
% Under this setup, this speed has evident practical costs, determining the amount of incoming samples that are missed by the model while it was busy processing previous samples. %  will miss samples that where revealed while it  in terms of the amount of samples the model misses when  observed by the model.
% allows accessing all the samples for adaptation, disregarding the fact that, when deployed, slower inferences at test time have a practical cost of missing data. % solely focuses on classification accuracy when the model pays no penalty, disregarding the impact of slower inferences at test time. % , thus assuming unlimited computational budget at inference.
% This evaluation hence putting more efficient approaches at a disadvantage.
% This idealistic assumption on computational budget stands in stark contrast to the realistic setting that TTA methods target, whereby DNNs are deployed in unseen domains and must  quickly adapt. Thus, reduced inference speed has particular consequences f
For TTA methods that adapt using test data, % for adaptation, 
missing samples has a direct effect on the method's accuracy, as it will have fewer samples for adaptation. %  to. % inference speed has a direct effect on the amount of data the model can leverage to adapt.
That is, the slower the TTA method, the fewer samples it can leverage for adapting to the distribution shift.
% In practical terms, this assumption amounts to presuming control over the frequency at which data arrives during testing.
% This supposition is untenable in the real world, where events unfold in the environment, and are presented to the model as a data \textit{stream} whose rate that is independent from the model's processing capacity.
Thus, the current protocol for evaluating TTA methods %which disregard inference speed, 
is not suitable for assessing their efficacy 
% and applicability 
in
% of these methods 
real-world deployment. %  when data, at test time, arrive independently from the speed of the deployed model.

In this work, we propose a novel \textit{realistic} evaluation protocol that factors in inference speed to assess the real-world applicability of TTA methods. % , and thus better .
% Our protocol accounts for the computational requirements for TTA methods in an effort to mimic real-world scenarios.
% Inspired by realistic evaluations in Continual Learning settings~\cite{ghunaim2023real}, we mimic realistic scenarios by exposing all TTA methods, regardless of their inference speed, to a constant data stream to be processed.
Our evaluation protocol is inspired by Online Learning~\cite{cloc, shalev2012online} and mimics real-world scenarios by exposing all TTA methods %regardless of their inference speed, 
to a constant-speed stream of data. % to be processed.
In this setting, the performance of slow TTA methods is intrinsically penalized, as the time spent adapting to a sample may lead to dropped samples that could have been useful for adaptation.
% We mimic such scenarios by penalizing expensive TTA methods and allows all methods similar \shyma{not similar but maybe "more realistic"?} computational budgets to adapt to a given stream of data.
% Specifically, our protocol dictates that if method $\mathcal A$ is $n$ times as slow as method $\mathcal B$, then method $\mathcal A$ may only use every $n^{\text{th}}$ batch for adapting to the stream, while $\mathcal B$ is allowed to adapt to every batch.
Specifically, our protocol dictates that if a method $g_{\text{slow}}$ is $k$ times slower than the stream, then %method $g_1$ 
it may only use every $k^{\text{th}}$ sample for adaptation. %to the stream, 
In contrast, a method $g_{\text{fast}}$ that is as fast as the stream is allowed to adapt to every sample. 
Figure \ref{fig:pull_fig} shows the effect of evaluating several methods under our proposed protocol, where slower methods (\emph{e.g.}, SAR~\cite{sar}) are penalized and faster but simpler methods become better alternatives~(\emph{e.g.}, SHOT~\cite{shot} and AdaBN~\cite{adabn}).

% Since several applications require the model's predictions at each given sample, we allow all methods to conduct a regular non-adaptive inference for all missed samples.
% By doing so, we attain a fair assessment of the effectiveness of TTA methods under identical % \shyma{Same thing should we say realistic vs similar} 
% computational budgets, unlike previous evaluations that grant unlimited budgets~\cite{}.
% We then apply this protocol to benchmark several TTA methods on various datasets, and provide a fair assessment of their performance subject to the realistic consequences of slowing inference speeds.
We apply our proposed evaluation protocol to benchmark several TTA methods on multiple datasets, and provide a fair assessment of their performance subject to the realistic consequences of slower inference speeds. 
Our experimental results highlight the importance of developing TTA methods that adapt to distribution shifts with minimal impact on inference speed. 
Our contributions are two-fold:
% Our experimental results highlight the real-world importance of developing TTA methods that adapt to distribution shifts with minimal impacts on inference speed. % , rendering them practical for real-world use.
% In summary, our contributions are twofold:
% \noindent\textbf{(i)} 
\begin{enumerate}
    \item We propose a realistic evaluation protocol for TTA methods that penalizes slower methods by providing them with fewer samples for adaptation. Our approach is 
    % simple yet 
    effective at assessing 
    % the efficacy of  
    TTA methods' efficacy in scenarios where data arrives as a constant-speed stream.

\item  Following our proposed protocol, we provide a comprehensive experimental analysis of 15 TTA methods evaluated on 3 large-scale datasets under 3 different evaluation scenarios.
These scenarios consider adaptation to a single domain and continual adaptation to several domains. 
Our analysis shows that, when inference speed is accounted for, simple (but faster) approaches can benefit from adapting to more data, and thus outperform more sophisticated (but slower) methods.
Figure~\ref{fig:pull_fig} demonstrates this for four TTA methods.
We hope our evaluation scheme inspires future TTA methods to consider inference speed as a critical dimension that affects their real-world performance.
\end{enumerate}

\section{Related Work}
\begin{figure*}
    \centering
    \includegraphics[width=0.99\linewidth]{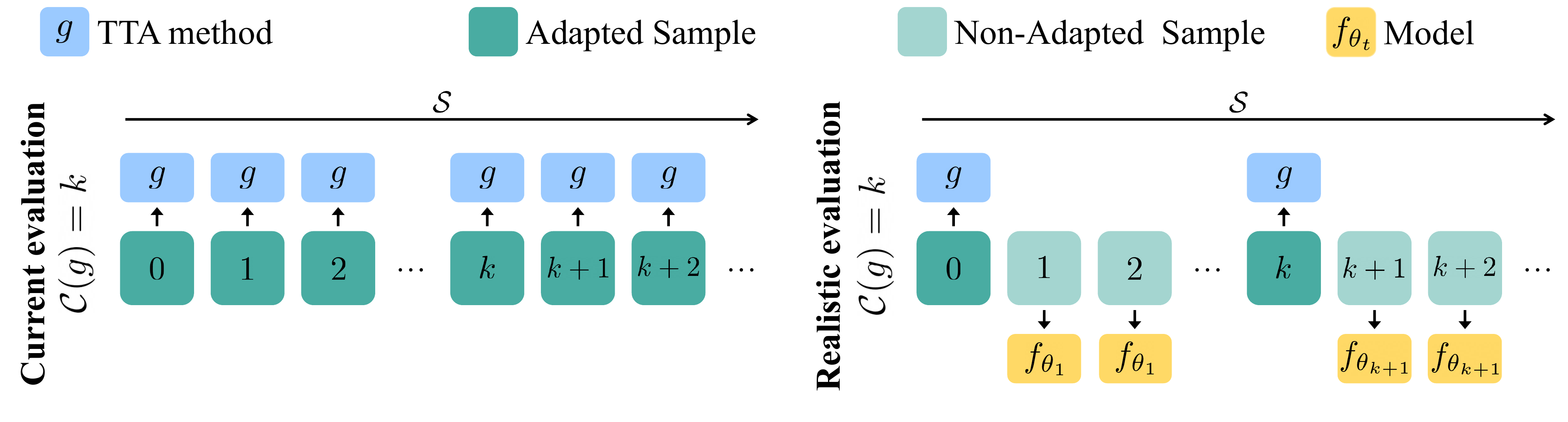}
    \caption{
    \textbf{Inference under the current and realistic evaluation protocols.} 
    The current evaluation setting (left) assumes that the incoming batches of stream $\mathcal{S}$ can wait until the adaptation process of a TTA method $g$ finishes. 
    This assumption is untenable in a real-time deployment scenario. 
    Our proposed realistic evaluation (right) simulates a more realistic scenario where $\mathcal S$ reveals data at a constant speed. 
    % the deployment of two models in parallel. 
    % The first model $g$ that is equipped with adaptation, and a second model $f_{\theta}$ without adaptation. 
    % $f_{\theta}$ can be the original base model that was never adapted or the most recent adapted model of $g$.
    In this setup, slower TTA methods will adapt to a smaller portion of the stream.
    The remaining part of the stream will be predicted without adaptation by employing the most recent adapted model.
    % We choose to use the most recent adapted model, which is arguably better than the base model. 
    We refer to the most recent adapted model as $f_{\theta_{t+1}}$, with $t$ denoting the time when the last sample was adapted to by $g$. 
    When $g$ is still adapting to a sample, the incoming sample is fed to $f_{\theta_{t+1}}$ to produce predictions.
}

    \label{fig:pipeline}
\end{figure*}

\paragraph{Test Time Adaptation.}
The Test Time Adaptation~(TTA) setup relaxes the ``i.i.d" assumption between the training and testing distributions~\cite{ttt, lame}.
This relaxation is usually attained through a lifelong learning scheme on all received unlabeled data~\cite{chen2022contrastive, gongnote}.
Earlier approaches such as TTT~\cite{ttt} and TTT++~\cite{liu2021ttt++}, among others~\cite{torralba2011unbiased, tzeng2017adversarial}, include a self-supervised loss~\cite{gidaris2018unsupervised} during training, which can then provide an error signal during adaptation.
Despite their effectiveness, such approaches assume having control over how the model is trained.
% In this work, we propose an online evaluation protocol that simulates exposing TTA methods to constant-speed stream of data.

\noindent\textbf{Fully Test Time Adaptation.}
% \subsection{Fully TTA Methods}
Fully TTA methods are a subtype of TTA method that adapts at test time by modifying the model's parameters~\cite{shot, pl, actmad, mancini2018kitting, cfa} or its input~\cite{dda} by using the incoming unlabeled data. % available during testing.
Fully TTA methods are practical, as they avoid assumptions on the training phase of a given model~\cite{tent, dda, t3a}.
The first of these approaches adjusts the statistics of the Batch Normalization~(BN) layers~\cite{dua, bnadaptation, adabn}.
For example, BN-adaptation~\cite{bnadaptation} leverages the statistics of the source data as a prior and infers the statistics for every received sample. 
On the other hand, AdaBN \cite{adabn} discards the statistics of the source domain and uses the statistics computed on the target domain. 
% These methods are computationally light, and their performance is expected to be preserved under online evaluation protocol.
In line with light TTA methods, LAME \cite{lame} proposes to only adapt the model's output by finding the latent assignments that optimize a manifold-regularized likelihood of the data.
In this work, we found that such efficient methods preserve their accuracy under our proposed evaluation.
While fully TTA methods have been studied in the context of adversarial domain shifts~\cite{alfarra2022combating,croce2022evaluating, perez2021enhancing}, in this work we focus on the context of natural shifts such as realistic image corruptions~\cite{imagenetc, 3dcc}.

Another line of work aims at adapting to distribution shifts by minimizing entropy. %  minimization.
For instance, SHOT \cite{shot} adapts the feature extractor to minimize the entropy of individual predictions; while maximizing the entropy of the predicted classes. %  in the received batch.
TENT~\cite{tent} updates the learnable parameters of the BN layers to minimize the entropy of predictions.
EATA~\cite{eata} combines TENT with an active selection of reliable and non-redundant samples from the target domain and an anti-forgetting loss~\cite{ewc}. % with an active sampling scheme for 
Further, SAR~\cite{sar} equips TENT with an active sampling scheme that filters samples with noisy gradients. %  when minimizing test entropy for a better life-long adaptation. 

Other works use data-augmentation at test time~\cite{tta}.
For example, MEMO~\cite{memo} adapts model parameters to minimize the entropy over a sample and multiple augmentations of it. %  the same test data point.
CoTTA \cite{cotta} uses augmentations to generate reliable pseudo-labels % from a teacher model 
and then peform distillation. % scheme.
Finally, DDA \cite{dda} proposes to leverage a diffusion model~\cite{ho2020denoising} to restore corrupted inputs back to the source data distribution. 
These methods require multiple forward passes through the network or a diffusion model, leading to slower inference speeds. % per a single data point.

\section{Methodology}\label{sec:method}
In this section, we present our proposed Realistic TTA evaluation protocol. 
We first describe the current TTA evaluation protocol and its limitations 
% in Section~\ref{sec:test time adaptation}. 
Then, 
% in Section~\ref{sec:real_time_tta}, 
we introduce our Realistic TTA evaluation protocol, which addresses the shortcomings of the offline protocol.

% This section introduces the proposed Online TTA evaluation protocol. 
% We first describe in Sec.~\ref{sec:test time adaptation} the current (offline) TTA evaluation protocol commonly used, and discuss its issues. 
% We then introduce our Online TTA evaluation protocol in Sec.~\ref{sec:real_time_tta}, which addresses such shortcomings. 

% The current offline TTA evaluation protocal neglects the consequences of exposing methods to a \textit{constant} stream of data, %  a method's inference speed has on its own performance.
% and thus provides an unrealistic assessment of their performance when deployed.
% Our protocol relies on the insight that, if the stream is constant, then a method's inference speed directly impacts its own performance. Slow methods will miss samples from the stream, and thus will adapt to fewer samples.
% % improves this assessment by noticing that a method's inference speed directly impacts the amount of samples the method adapts on. % impacts stream of data, whereby the method's inference speed directly impacts the amount of samples the method can access for adaptation. % as accounting  problem on a stream of data.
% As such, our protocol penalizes slow methods in a realistic manner, and thus provides an assessment of the real-world applicability of TTA methods. Next, we give a brief background on TTA, and then elaborate on our proposed protocol. % to assess the real-world applicability of TTA methods.
% % We then propose our adjusted real-time evaluation scheme.

\subsection{Current Protocol}\label{sec:test time adaptation}
% \red{We need to cite CLOC somewhere here as we adapted the interaction from the online learning there}

% Introductory paragraph

% or alternatively, test time training, 
TTA considers the practical setup, in which trained models are deployed in a target domain that exhibits distribution shifts to which they must adapt.
Let $f_{{\theta}}: \mathcal X \rightarrow \mathcal Y$ be a classifier, parameterized by ${\theta}$, that predicts the label $y\in \mathcal Y$ for a given input $x\in\mathcal X$. 
Before test time, $f_{{\theta}}$ is assumed to have been trained on the dataset $\mathcal D_{\text{train}} \subset \mathcal X \times \mathcal Y$.
At test time, \emph{i.e.} when executing TTA, $f_{{\theta}}$ is presented with a stream of data $\mathcal S$, sampled from $\mathcal{X}$, with potentially multiple distribution shifts w.r.t. $\mathcal D_\text{train}$.
Under this setup, a TTA method is a function % $g(\theta, x): \Theta \times \mathcal X \rightarrow \Theta \times \mathcal X$ 
$g(\theta, x)$ that sequentially adapts the model's parameters $\theta$ and/or the input $x$ to enhance the performance under distributions shifts. %  to the data stream $\mathcal S$. % distribution shift.
Currently, TTA methods are evaluated in an offline setting. 

Formally, the \textit{Current TTA evaluation protocol} simulates the interaction between the stream $\mathcal S$ and the TTA method $g$, at each time step $t \in \{0,1,\dots,\infty\}$, as follows: %  At each test time step $t\in\{1, 2, \dots, \infty\}$, :
\begin{comment}
\begin{enumerate}
    \item $\mathcal S$ reveals a batch of images $\{x_i^t\}_{i=1}^{n_t}\in\mathcal X$.
    \item $f_{\theta_t}$ generates the predictions $\{\hat{y}_i^t\}_{i=1}^{n_t}$ for $\{x_i^t\}_{i=1}^{n_t}$.
    \item The TTA algorithm $g$ adjusts the predictions by adapting $\theta_t$ and/or $\{x_i^t\}_{i=1}^{n_t}$ to $\hat{\theta}_t$ and $\{\hat{x}_i^t\}_{i=1}^{n_t}$. Model parameters are updated with $\theta_t \leftarrow \alpha \theta_t + (1-\alpha)\hat{\theta}_t$.
\end{enumerate}
\end{comment}
\begin{enumerate}[label=\textcolor{black}{\textbf{Curr.\arabic*}}, leftmargin=*]
    \item\label{reveal} $\mathcal S$ reveals a sample $x_t$. % batch of samples $X_t = \{x_t^i\}_{i=1}^{n_t}$. % , the batch of samples corresponding to time step $t$. % , \:x_i^t\in\mathcal X$.
    % \item $f_{\theta_t}$ generates predictions $\hat{Y}_t = \{\hat{y}_t^i\}_{i=1}^{n_t}$ for $X_t$.\label{predict} % $\{x_i^t\}_{i=1}^{N}$.
    \item\label{adapt} $g$ adapts $x_t$ to $\hat{x}_t$, $\theta_t$ to $\hat{\theta}_t$, generates prediction $\hat{y}_t$, and updates parameters $\theta_{t+1} = \alpha \theta_t + (1-\alpha)\hat{\theta}_t$.\footnote{Note that some methods abstain from adapting either $x_t$ or $\theta_t$.} % strict?
    % $g$ adapts to the batch by adjusting the parameters, and/or the inputs $X^t$.
    % The parameters are adjusted via $\theta \leftarrow \alpha \theta + (1-\alpha)\hat{\theta}$, where $\hat{\theta}$ are parameter updates generated by $g$.
    % The predictions $\hat{Y}^t$ may be recomputed according to these adjustments. %  adjusts the predictions by adapting $\theta_t$ and/or $\{x_i^t\}_{i=1}^{N}$ to $\hat{\theta}_t$ and $\{\hat{x}_i^t\}_{i=1}^{N}$. 
    % Model parameters are updated with $\theta_t \leftarrow \alpha \theta_t + (1-\alpha)\hat{\theta}_t$.
    % \item Set $t \leftarrow t + 1$, and continue to \ref{reveal}.
\end{enumerate}

% \juan{I'm currently not sure we should include this notation having to do with the explicit form of parameter updates. We never use it again, and it's not really related to what we do, no? I think the only important part of the next sentences is \textit{``Note that all current TTA methods can be described by this sequence of events.''} (and even that is not \textit{that} important.)}
% The parameter adaptations can be described by $\theta \leftarrow \alpha \theta + (1-\alpha)\hat{\theta}$, where $\hat{\theta}$ are parameter updates generated by $g$.

Note that all existing TTA methods can be modeled using this framework.
For example, TENT~\cite{tent} adapts network parameters to minimize entropy with $\alpha=0$, while leaving inputs  unchanged, \emph{i.e.} $\hat{x}_t = x_t$ and $\theta_{t+1} = \hat{\theta}_t$. 
DDA~\cite{dda} adapts inputs via a diffusion process while preserving network parameters with $\alpha=1$, \emph{i.e.} $\hat{x}_t = \hat{x}_t$ and $\theta_{t+1} = {\theta}_t$.
CoTTA~\cite{cotta} applies knowledge distillation, and updates network parameters with an exponential moving average, \emph{i.e.} setting $0<\alpha<1$.

\noindent\textbf{Shortcomings of the Current TTA protocol.}
In the current protocol, the performance of a TTA method $g$ is measured by comparing the ground truth labels $y_t$ with the predictions after adaptation $\hat y_t$.
An evaluation based only on this measure implicitly assumes that the % speed of the data stream is dependent on the speed of $g$. 
% Under this assumption, the 
stream is not constant speed, but rather waits for $g$ to adapt to $x_t$ ~(\ref{adapt}) before revealing the next batch $x_{t+1}$~(\ref{reveal}).
% $g$ adapts to \textit{every} sample from the stream $\mathcal S$
% \footnote{Recently, efficient TTA methods~\cite{sar, eata} emerged that adapt to a subset of the stream. However, such methods need to update to the full stream in the worst-case scenario.}, 
Figure~\ref{fig:pipeline} provides an illustration of this situation.
% That is, $\mathcal S$ will not reveal any new images (step 1) until $g$ adapts to the previous batch (step 3).
This assumption results in the offline protocol favoring slower TTA methods, as the method's performance is agnostic to its inference speed. However, in practical applications where the test data arrives at a constant speed, the offline protocol is  not suitable for assessing a method's performance. %  in practical scenarios.
Next, we propose a remedy for this shortcoming.
% This assumption results in an evaluation that favors slow methods, since the method's performance is independent of its speed. 
% In online applications where the test data arrives at a constant speed, the offline evaluation may be inconsistent with the performance the method would have in practice.
% Furthermore, the stream is usually assumed to contain a single distribution shift (\emph{e.g.} a single corruption) where classes are mixed within batches~\cite{tent, cotta}, as outlined in previous works~\cite{sar, dda}.\red{ALEJO: Is this True for every setup? what about the case of a single model deployed in several locations? this might cause several shifts within a batch, doesn't it?}

% Hence, one should fairly compare TTA methods when evaluated on real-time streams to assess their effectiveness when taking their computational complexity into account.
% Further, some TTA methods employ cumulative updates during adaptation (\emph{i.e.} setting $\alpha < 1$) that works well when adapting to a single distribution shift~\cite{tent, eata, cotta}. 
% Such methods fail when deployed on a continuous lifelong adaptation~\cite{sar}.
% Furthermore, several TTA approaches are extremely sensitive to the batch size that they are tested against~\cite{ttac, adabn}.
% Other TTA approaches partially alleviate the aforementioned methods by designing input-dependent adaptation approaches~\cite{memo, dda}.

% \paragraph{Fixing the evaluation protocol.}

\subsection{Realistic Online Evaluation Protocol}\label{sec:real_time_tta}
We propose a realistic evaluation of TTA methods that \emph{explicitly} considers the relation between the speed of the method and the speed at which the stream reveals new data.
This setup is more realistic, as it intrinsically penalizes the performance of slower TTA methods: \textit{long times spent in adaptation result in fewer samples to adapt to}.
% Thus, we address the current limitations in the evaluation protocol by analyzing a real-time evaluation of TTA methods that penalizes expensive adaptation methods.
% Further, we consider different evaluation scenarios (\emph{e.g.} adapting to a single or a continuous distribution shift) under our real-time (\red{alejo: time-constrained setup sounds safer, real-time implies other stuff we are not considering like synchronization, for instance} setup.

A crucial aspect of our realistic TTA protocol is accounting for the implications of simulating a constant speed data stream $\mathcal S$. % , which reveals data at a rate that is independent of the speed at which the TTA method processes such data.
% Key to our online TTA protocol is considering the consequences of simulating a constant stream $\mathcal S$, whose rate for revealing batches, denoted by $r$, is independent of the speed of the method being evaluated. %  the adaptation of a given TTA method is.
For instance, consider a stream $\mathcal S$ that reveals data at a constant rate $r$ samples per second.
If a method $g_\text{fast}$ adapts to samples at speed $r$, then $g_\text{fast}$ will be able to adapt to every sample. 
On the other hand, if $g_\text{slow}$ adapts to samples at a speed $\nicefrac{r}{2}$, then $g_\text{slow}$ will skip every other sample.
% half of the samples and thus will only adapt to the other half. %  in adaptation to only half of the total batches.
% For instance, if a method $g_\text{fast}$ adapts to data as fast as $\mathcal S$ reveals data, $g_\text{fast}$ will be able to adapt to every batch.
% In contrast, if a method $g_\text{slow}$ adapts to data twice as slow as $\mathcal S$ reveals data, then $g_\text{slow}$ skips half the batches, and only adapts to  \textit{every other} batch (step 3).
We formalize the notion of the relation between the speed of the stream and the speed of a method $g$ as the ``relative adaptation speed of $g$''. % , $\mathcal C(g) = \lceil \nicefrac{R(g)}{r} \rceil \in \mathbb N$.
% To incorporate this adjustment into the interaction between the stream $\mathcal S$ and a TTA method $g$, we define the relative adaptation complexity $\mathcal C(g) \in \mathbb N$.
% This quantity represents the relative speed of adapting to a received batch from $\mathcal S$ using $g$ to the rate at which $\mathcal S$ reveals new batches.
This quantity, denoted by $\mathcal C(g) \in \mathbb N$, is simply the integer ratio of the speed of $\mathcal{S}$ to the speed of $g$. % 's speed the time consumed by $g$ for adapting to a batch, and the time at which $\mathcal S$ reveals a batch.
For instance, in the previous example, $\mathcal C(g_\text{fast})=1$, meaning $g_\text{fast}$ adjusts as fast as $\mathcal S$ reveals data,
% On the other hand, a 
while $\mathcal C(g_\text{slow})=2$, indicating $\mathcal S$ reveals its second batch while $g_\text{slow}$ is still adapting to the first one. % , implying $g$ will output a non-adapted prediction on every other batch using the latest updated model~(step 2).

Without loss of generality, we assume that $f_{\theta}$ runs in real-time, \emph{i.e.} that its speed is equal to $r$, and thus $\mathcal C(f_{\theta}) = 1$.
This assumption allows us to suppose that the samples that are not processed by $g$ can be processed by $f_{\theta}$.
% Note that a TTA method skipping batches implies that the adapted model does not output predictions for such 
% However, since $g_2$ is in the deployment phase, $g_2$ needs to predict every revealed batch from $\mathcal S$ (step 2).
% \red{Maybe we need a proper definition here}
% Without loss of generality, normalize the stream speed with respect to the fastest inference, \emph{i.e.} the Naive forward pass without any adaptation.
Under this setup, we define our realistic protocol by introducing the relative adaptation speed $\mathcal C(g)$ into the offline protocol.
In particular, we simulate $g$'s availability by \textit{conditionally} performing the adaptation step (\ref{adapt}), depending on $\mathcal C(g)$.
In this manner, $g$ is only permitted to adapt when its previous adaptation step has finished.
% That is, $g$ is only allowed to adapt if its previous adaptation would have finished.
Formally, the \textit{realistic TTA evaluation protocol} simulates the interaction between the constant speed stream $\mathcal{S}$ and the TTA method $g$, % (with $\mathcal C(g) = k$), 
at each time step $t \in \{0, 1, \dots, \infty\}$, as follows:
% With that, one can define the relative computational overhead of a given TTA method $\mathcal A$;  $\mathcal O(\mathcal A)$, as the ratio between its computation and the computation required to do a naive forward pass.
% Therefore, $\mathcal O(\mathcal A) = 1$ for a method that is as fast as the stream, and $\mathcal O(\mathcal B) = 2$ for a twice as slow method.
% We need to define the delay for a method. Also, we need to explain how are we going to calculate it. Suppose that a TTA method has a delay of $k$, then the realistic setup of TTA should follow:
\begin{comment}
\begin{enumerate}
    \item $\mathcal S$ reveals a batch of images $\{x_i^t\}_{i=1}^{n_t}\in\mathcal X$.
    \item $f_{\theta_t}$ generates the predictions $\{\hat{y}_i^t\}_{i=1}^{n_t}$ for $\{x_i^t\}_{i=1}^{n_t}$.
    \item If $\mathrm{mod}(t, k)=0$, then a TTA algorithm $g$ adjusts the predictions by adapting $\theta_t$ and $\{x_i^t\}_{i=1}^{n_t}$ to $\hat{\theta}_t$ and $\{\hat{x}_i^t\}_{i=1}^{n_t}$. Further, update $\theta_t \leftarrow \alpha \theta_t + (1-\alpha)\hat{\theta}_t$
\end{enumerate}

\item\label{reveal} $\mathcal S$ reveals a batch of samples $X_t = \{x_t^i\}_{i=1}^{n_t}$. % , the batch of samples corresponding to time step $t$. % , \:x_i^t\in\mathcal X$.
    \item $f_{\theta_t}$ generates predictions $\hat{Y}_t = \{\hat{y}_t^i\}_{i=1}^{n_t}$ for $X_t$. % $\{x_i^t\}_{i=1}^{N}$.
    \item\label{adapt} $g$ adjusts $\hat{Y}_t$ by adapting $\theta_t$ to $\hat{\theta}_t$ and $X_t$ to $\hat{X}_t$. 
    Further, define $\theta_{t+1} = \alpha \theta_t + (1-\alpha)\hat{\theta}_t$.
\end{comment}
\begin{enumerate}[label=\textcolor{black}{\textbf{RTTA \arabic*}}, leftmargin=*]
    \item\label{reveal_online} $\mathcal S$ reveals a sample $x_t$. % batch of samples $X_t = \{x_t^i\}_{i=1}^{n_t}$. % , \:x_i^t\in\mathcal X$.
    % \item\label{predict_online} $f_{\theta_t}$ generates predictions $\hat{Y}_t = \{\hat{y}_t^i\}_{i=1}^{n_t}$ for $X_t$. % $\{x_i^t\}_{i=1}^{N}$.
    \item\label{adapt_online} \underline{If $\:\:\left(t\:\:\mathrm{mod}\:\:\mathcal{C}(g)\right) = 0,\:$ then} % \item\label{adapt_online} \underline{If $\mathrm{mod}(t, k)=0$ then} 
    $g$ adapts $x_t$ to $\hat{x}_t$, $\theta_t$ to $\hat{\theta}_t$, generates a prediction $\hat{y}_t$, and updates parameters via $\theta_{t+1} \leftarrow \alpha \theta_t + (1-\alpha)\hat{\theta}_t$.
    \\\underline{Otherwise}, $f_{\theta_t}$ generates a prediction $\hat{y}_t$.
\end{enumerate}
Here, ``$\mathrm{mod}$" represents the modulo operation. 
The above protocol assesses the performance of TTA methods by factoring in their speed.
As such, faster methods are granted more adaptation steps and, conversely, slower methods are granted fewer % of $\mathcal S$~
(see Figure~\ref{fig:pipeline}). 
Note that explicitly modeling the relative adaptation speeds allows us to evaluate TTA methods under different adaptation speeds by setting $\mathcal{C}(g)$ to arbitrary values. 
For instance, note that our realistic protocol recovers the original offline protocol by setting $\mathcal C(g) = 1$ for all methods. Next, we explain the calculation of $\mathcal{C}(g)$ for our realistic protocol.

% \juan{(ETA or EATA? Somebody who knows the literature please clarify this.)}
\paragraph{Online computation of $\mathcal C(g)$.}
In practice, estimating the relative adaptation speed $\mathcal C(g)$ % , fundamental to our online evaluation, 
% requires accounting for its variation across time. 
can be a noisy process.
The noise in this estimation essentially comes  from two factors: hardware and input dependence.
Hardware-induced noise applies to all methods, while input dependence applies to methods like ETA~\cite{eata} 
 which, upon receiving an input, may optionally abstain from adapting to it.
This noise means that $\mathcal C(g)$ potentially varies across iterations.
% To evaluate TTA methods using our proposed realistic evaluation, one should carefully handle their relative adaptation complexity.
% We normalize the stream speed by the speed of the forward pass of the non-adapted model~$f_\theta$, \emph{i.e.} that $\mathcal{C}(f_\theta) = 1$.
% On one hand, some TTA methods induce a fixed amount of additional computation to adapt to any given batch.
% For example, we experimentally find that $\mathcal C(g_{\text{TENT}}) = 3$  for TENT~\cite{tent}, which correlates with TENT's single-step optimization for % of entropy minimization to 
% updating the % learnable parameters of 
% batch normalization layers.
% For instance, our experiments show $\mathcal C(g_{\text{tent}}) = 3$ since the cost of conducting a backward pass is $2\times$ the cost of the forward pass. \juan{I know this is usually a rule-of-thumb (although I had heard of this in terms of memory consumption, not time), but do we have a ``citable'' source for this?}
% On the other hand, other TTA methods add a variable cost depending on the received batch.
% For instance, ETA~\cite{eata} rejects some samples for adapting the batch normalization layers, resulting in a variable
% model's  parameters of the model, %  on samples with high entropy.
% $\mathcal C(g_\text{ETA})$ across batches.

Our protocol accounts for this variability by conducting an online computation of $\mathcal C(g)$ on each revealed input. %  for each method.
That is, instead of using a fixed value of $\mathcal{C}(g)$ at each iteration~$t$, our protocol rather uses $\mathcal{C}\left(g(x_t)\right)$.
Formally, if we let $R\left(g(x)\right)$ denote the speed at which $g$ processes $x$, then the relative adaptation speed of $g$ at $x$ is defined as $\mathcal{C}\left(g(x_t)\right) = \lceil \nicefrac{r}{R\left(g(x)\right)} \rceil$, where the ceiling function accounts for the stream's discrete-time nature.
Note that since we assumed $\mathcal C(f_\theta) = 1$, then $R\left(f_\theta(x)\right) = r$.
% We compare the computational complexity to conduct the $g(x)$, denoted by $\mathcal L(g(x))$, to  $\mathcal L(f_\theta(x))$ in terms of run-time.
% Since $\mathcal C(f_\theta) = 1$, then we calculate $\mathcal C(g) = \lceil \nicefrac{\mathcal L(g(x))}{\mathcal L(f_\theta(x))} \rceil$.
% Note that this online computation of $\mathcal{C}\left(g(x_t)\right)$ is only performed at step \ref{adapt_online}, when $g$ adapts to the batch from the stream.
% By doing so, we accommodate for both TTA methods that incur either fixed or variable computational costs for each batch.
We report the empirical behavior of this online computation of $\mathcal{C}\left(g(x_t)\right)$ for various TTA methods in Table~\ref{tab:cg_methods}, and leave the rest of the methods and the computation details to the Appendix.
Next, we leverage our Realistic TTA protocol to conduct a comprehensive empirical study of several TTA methods. %  when evaluated in a realsetup.

\begin{table}[t]
\caption{\label{tab:cg_methods}\textbf{Average $\mathcal C(g(x_t))$.}
We report the average relative adaptation speed $\mathcal C(g)$ for 5 TTA methods.
The higher $\mathcal C(g)$ is, the smaller the portion of data to which $g$ adapts is.}
\center
% \footnotesize
\vspace{-5pt}
% \resizebox{\linewidth}{!}{
\setlength{\tabcolsep}{2pt}
\begin{tabular}{l|c|c|c|c|c}
\toprule
Method & AdaBN & TENT & TTAC-NQ & MEMO & DDA\\ 
\midrule 
$\mathcal C(g)$ & 1 & 3 & 12 & 54  & 810 \\
\bottomrule
\end{tabular}\vspace{-0.5cm}
% }
\end{table}

\section{Experiments}
% \usepackage{multirow}
% \usepackage[table.xcdraw]{xcolor}
% If you use a beamer only pass "xcolor=table" option. i.e. \documentclass[xcolor=table]{beamer}
\begin{table*}[]
\caption{\label{tab:imagenetc_episodic} \textbf{Episodic Error Rate on ImageNet-C.}
We report the error rate of different TTA methods on ImageNet-C benchmark under both the realistic and the current setup. A lower error rate indicates a better TTA method. The highlighted numbers indicate a better performance per method across setups. Episodic means the model will adapt to one corruption at a time. The model is reset back to the base model when moving to the next corruption. The current setup is merely the reproduction of every method. The first sub-table corresponds to methods that do not incur any or few extra computations, \emph{i.e.} $\mathcal{C}(g) = 1$. We show that methods generally perform worse in the realistic setup. The more computationally complex the TTA method is, the less data it will adapt to, and the worse is its performance. 
% \red{we need to devise a better grouping like the one we did in the related work, TENT/SAR/EATA are entropy minimization methods, MEMO/CoTTA use consistency over augmentations, AdaBN/BN adapt batch norm statistics etc.}
}\vspace{-0.1cm}
\resizebox{\textwidth}{!}{
% \usepackage{multirow}
% \usepackage[table.xcdraw]{xcolor}
% If you use beamer only pass "xcolor=table" option. i.e. \documentclass[xcolor=table]{beamer}
\begin{tabular}{l|c|ccc|cccc|cccc|cccc|c|c}
\toprule
\rowcolor[HTML]{FFFFFF} 
\cellcolor[HTML]{FFFFFF}                          & \cellcolor[HTML]{FFFFFF}                        & \multicolumn{3}{c|}{\cellcolor[HTML]{FFFFFF}Noise} & \multicolumn{4}{c|}{\cellcolor[HTML]{FFFFFF}Blur}                           & \multicolumn{4}{c|}{\cellcolor[HTML]{FFFFFF}Weather} & \multicolumn{4}{c|}{\cellcolor[HTML]{FFFFFF}Digital} & \cellcolor[HTML]{FFFFFF}                       \\
\rowcolor[HTML]{FFFFFF} 
\multirow{-2}{*}{\cellcolor[HTML]{FFFFFF}Method}  & \multirow{-2}{*}{\cellcolor[HTML]{FFFFFF}Realistic} & gauss.                         & shot   & impul.  & defoc.                       & glass & motion                       & zoom & snow       & frost       & fog        & brigh.      & contr.      & elast.      & pixel.      & jpeg      & \multirow{-2}{*}{\cellcolor[HTML]{FFFFFF}Avg.} & \multirow{-2}{*}{\cellcolor[HTML]{FFFFFF}$\Delta$}\\
\midrule
\rowcolor[HTML]{FFFFFF} 
Source                                            & \cmark                                          & 97.8                           & 97.1   & 98.1    & 82.1                         & 90.2  & 85.2                         & 77.5 & 83.1       & 76.7        & 75.6       & 41.1        & 94.6        & 83.0        & 79.4        & 68.4      & 82.0 & -                                           \\%\midrule \midrule
\rowcolor[HTML]{FFFFFF} 
AdaBN                                           & \cmark                                          & 84.9                           & 84.3   & 84.3    & 85.0                         & 84.7  & 73.6                         & 61.1 & 65.8       & 66.9        & 52.1       & 34.8        & 83.3        & 56.1        & 51.1        & 60.3      & 68.5 & -                                          \\%\midrule
\rowcolor[HTML]{FFFFFF} 
LAME                                             & \cmark                                          & 98.3 & 97.6 & 98.6 & 82.4 & 90.9 & 86.1 & 78.1 & 84.5 & 77.5 & 77.3 & 41.4 & 94.8 & 84.8 & 80.0 & 68.9  & 82.7 & -                                        \\%\midrule
\rowcolor[HTML]{FFFFFF} 
BN                                              & \cmark                                          & 84.6                           & 83.9   & 83.8    & 80.1                         & 80.2  & 71.7                         & 60.4 & 65.4       & 65.2        & 51.6       & 34.6        & 76.3        & 54.4        & 49.7        & 59.2      & 66.7 & -                                          \\
\midrule
\midrule
\rowcolor[HTML]{FFFFFF} 
\cellcolor[HTML]{FFFFFF}                          & \cellcolor[HTML]{FFFFFF}\xmark                 & \cellcolor[HTML]{e9f5f9}73.4   & 70.2   & 73.0    & 76.6                         & 75.5  & \cellcolor[HTML]{e9f5f9}59.8 & 53.8 & 54.2       & 63.4        & 44.7       & 35.5        & 79.3        & 46.9        & 43.2        & 49.7      & 59.9                                            \\
\rowcolor[HTML]{e9f5f9} 
\multirow{-2}{*}{\cellcolor[HTML]{FFFFFF}SHOT}    & \cellcolor[HTML]{FFFFFF}\cmark                  & \cellcolor[HTML]{FFFFFF}73.6   & 69.0   & 71.1    & 74.6                         & 74.8  & \cellcolor[HTML]{FFFFFF}60.0 & 52.9 & 54.1       & 61.3        & 44.1       & 34.1        & 77.8        & 46.8        & 43.1        & 49.2      & 59.1 & \multirow{-2}{*}{\cellcolor[HTML]{FFFFFF} \textcolor{GTgreen}{(-0.8)}}                                  \\
\midrule
\rowcolor[HTML]{e9f5f9} 
\cellcolor[HTML]{FFFFFF}                          & \cellcolor[HTML]{FFFFFF}\xmark                 & 71.3                           & 69.4   & 70.2    & 72.0                         & 72.9  & 58.7                         & 50.7 & 52.8       & 58.8        & 42.7       & 32.7        & 73.3        & 45.5        & 41.5        & 47.7      & 57.3                                         \\
\rowcolor[HTML]{FFFFFF} 
\multirow{-2}{*}{\cellcolor[HTML]{FFFFFF}TENT}    & \cellcolor[HTML]{FFFFFF}\cmark                  & 75.7                           & 78.3   & 75.2    & 76.3                         & 77.3  & 64.6                         & 55.6 & 57.3       & 61.4        & 45.9       & 33.5        & 77.1        & 50.1        & 44.2        & 51.4      & 61.6 & \multirow{-2}{*}{\textcolor{red}{(+4.3)}}                                          \\
\midrule
\rowcolor[HTML]{e9f5f9} 
\cellcolor[HTML]{FFFFFF}                          & \cellcolor[HTML]{FFFFFF}\xmark                 & 69.5                           & 69.7   & 69.0    & 71.2                         & 71.7  & 58.1                         & 50.5 & 52.9       & 57.9        & 42.7       & 32.7        & 62.9        & 45.5        & 41.6        & 47.8      & 56.2                                         \\
\rowcolor[HTML]{FFFFFF} 
\multirow{-2}{*}{\cellcolor[HTML]{FFFFFF}SAR}     & \cellcolor[HTML]{FFFFFF}\cmark  
% & 84.8                           & 84.2   & 84.0    & 84.9                         & 84.4  & 73.4                         & 60.8 & 65.6       & 66.7        & 51.9       & 34.8        & 83.1        & 56.0        & 50.9        & 60.1      & 68.4 
&79.4	&78.5	&78.1	&79.9	&79.3	&67.5	&56.1	&60.5	&63.1	&47.4	&34.0	&75.3	&51.7	&46.6	&53.8	&63.4
& \multirow{-2}{*}{\textcolor{red}{(+7.2)}}                                          \\
\midrule
\rowcolor[HTML]{e9f5f9} 
\cellcolor[HTML]{FFFFFF}                          & \cellcolor[HTML]{FFFFFF}\xmark                 & 78.4                           & 77.8   & 77.2    & 80.5                         & 79.1  & 64.0                         & 53.3 & 57.8       & 60.7        & 44.1       & 32.9        & 73.1        & 48.6        & 42.3        & 52.6      & 61.5 & \cellcolor[HTML]{FFFFFF}                                         \\
\rowcolor[HTML]{FFFFFF} 
\multirow{-2}{*}{\cellcolor[HTML]{FFFFFF}CoTTA}   & \cellcolor[HTML]{FFFFFF}\cmark                  & 82.9                           & 81.6   & 81.9    & 87.4                         & 85.6  & 75.6                         & 61.1 & 63.1       & 64.9        & 49.9       & 34.8        & 91.2        & 54.0        & 48.8        & 56.6      & 68.0 & \multirow{-2}{*}{\textcolor{red}{(+6.5)}}                                           \\
\midrule\midrule
\rowcolor[HTML]{e9f5f9} 
\cellcolor[HTML]{FFFFFF}                          & \cellcolor[HTML]{FFFFFF}\xmark                 & 71.3                           & 70.3   & 70.8    & 82.1                         & 77.4  & 63.9                         & 53.9 & 49.9       & 55.5        & 43.9       & 32.8        & 81.4        & 43.7        & 41.1        & 46.7      & 59.0                                       \\
\rowcolor[HTML]{FFFFFF} 
\multirow{-2}{*}{\cellcolor[HTML]{FFFFFF}TTAC-NQ} & \cellcolor[HTML]{FFFFFF}\cmark                  & 79.4                           & 75.7   & 78.9    & 86.6                         & 86.2  & 77.1                         & 61.8 & 58.8       & 62.4        & 51.5       & 34.4        & 88.5        & 52.1        & 49.1        & 55.5      & 66.5  & \multirow{-2}{*}{\cellcolor[HTML]{FFFFFF}\textcolor{red}{(+7.5)}}                                         \\
\midrule
\rowcolor[HTML]{e9f5f9} 
\cellcolor[HTML]{FFFFFF}                          & \cellcolor[HTML]{FFFFFF}\xmark                 & 65.5                           & 62.4   & 63.5    & 66.6                         & 67.2  & 52.0                         & 47.3 & 48.2       & 54.1        & 39.9       & 32.1        & 55.0        & 42.3        & 39.2        & 44.8      & 52.0                                          \\
\rowcolor[HTML]{FFFFFF} 
\multirow{-2}{*}{\cellcolor[HTML]{FFFFFF}EATA}    & \cellcolor[HTML]{FFFFFF}\cmark                  & 69.3                           & 67.1   & 69.2    & 71.1                         & 71.7  & 57.5                         & 49.9 & 51.9       & 57.4        & 42.4       & 32.6        & 60.7        & 45.1        & 41.4        & 47.4      & 55.6  &  \multirow{-2}{*}{\cellcolor[HTML]{FFFFFF}\textcolor{red}{(+3.6)}}                                       \\
\midrule\midrule
\rowcolor[HTML]{e9f5f9} 
\cellcolor[HTML]{FFFFFF}                          & \cellcolor[HTML]{FFFFFF}\xmark                 & 92.5                           & 91.3   & 91.0    & \cellcolor[HTML]{FFFFFF}84.0 & 87.0  & 79.3                         & 72.4 & 74.6       & 71.3        & 67.9       & 39.0        & 89.0        & 76.2        & 67.0        & 62.4      & 76.3                                          \\
\rowcolor[HTML]{FFFFFF} 
\multirow{-2}{*}{\cellcolor[HTML]{FFFFFF}MEMO~}    & \cellcolor[HTML]{FFFFFF}\cmark                  & 97.7                           & 97.0   & 98.0    & \cellcolor[HTML]{e9f5f9}82.1 & 90.1  & 85.1                         & 77.4 & 83.0       & 76.6        & 75.4       & 41.0        & 94.5        & 82.9        & 79.2        & 68.2      & 81.9 & \multirow{-2}{*}{\cellcolor[HTML]{FFFFFF}\textcolor{red}{(+5.6)}}                                           \\
\midrule
\rowcolor[HTML]{e9f5f9} 
\cellcolor[HTML]{FFFFFF}                          & \cellcolor[HTML]{FFFFFF}\xmark                 & 58.6                           & 57.8   & 59.0    & \cellcolor[HTML]{FFFFFF}87.0 & 81.6  & 76.6                         & 65.9 & 67.9       & 66.7        & 64.0       & 40.0        & 92.2        & 52.2        & 46.6        & 49.9      & 64.4                                           \\
\rowcolor[HTML]{FFFFFF} 
\multirow{-2}{*}{\cellcolor[HTML]{FFFFFF}DDA}    & \cellcolor[HTML]{FFFFFF}\cmark                  & 97.8                           & 97.0   & 98.1    & \cellcolor[HTML]{e9f5f9}82.1 & 90.2  & 85.2                         & 77.5 & 83.1       & 76.7        & 75.6       & 41.1        & 94.6        & 83.0        & 79.4        & 68.3      & 82.0 & \multirow{-2}{*}{\cellcolor[HTML]{FFFFFF}\textcolor{red}{(+17.6)}}                                         \\ \midrule \bottomrule
\end{tabular}
% \vspace{-0.2cm}
}%As described in ~Sec.\ref{sec:episodic_eval}, we divide all methods  }
\end{table*}

% \subsection{Setup}\label{sec:setup}

We follow prior art~\cite{tent,sar,dda} and focus on the task of image classification.
In all our experiments, we assume that $f_\theta$ is a ResNet-50-BN\footnote{SAR demonstrated the superiority of using batch independent normalization layers under batch size of 1. We leave this ablation to the Appendix along with experiments on other architectures.}~\cite{he2016deep} trained on ImageNet~\cite{deng2009imagenet} (pretrained weights obtained from \texttt{torchvision}).
We further assume that the stream $\mathcal S$ reveals batches of size 64\footnote{This batch size is recommended by most baselines~\cite{tent,eata}}, except for MEMO~\cite{memo}, which predicts on single images to incentivize prediction consistency over an input and its augmentations.
% For all our experiments, we assume that the stream $\mathcal S$ reveals batches of size 64\footnote{We note that this is the batch size used on most baselines.} except for MEMO~\cite{memo} and DDA~\cite{dda} as they are data-dependent approaches.
%; we leave the ablations for other batch sizes in the appendix.
Regarding datasets, we follow earlier works~\cite{tent, sar, eata, dda, memo}, and thus evaluate on the ImageNet-C dataset~\cite{imagenetc} with a corruption level of~5 for all 15 corruptions.
We further extend our evaluation and consider CIFAR10-C, ImageNet-R~\cite{imagenetr}, and the more recent ImageNet-3DCC~\cite{3dcc}, which leverages depth estimates to construct more spatially-consistent corruptions.

Our experiments compare the performance of the baseline model $f_\theta$ %, trained solely on the source domain 
(without test time adaptation) %  (\emph{i.e.} classifying with $f_\theta$).
against 15 state-of-the-art TTA methods published in top-tier venues (\emph{e.g.},~CVPR, NeurIPS, and ICLR) between 2017 and 2023.
In particular, we consider: BN~\cite{bnadaptation} and AdaBN~\cite{adabn}, which adjust the statistics of the batch normalization layers; SHOT~\cite{shot} and SHOT-IM~\cite{shot}, which fine-tune the feature extractor to maximize mutual information; entropy minimization approaches such as TENT~\cite{tent}, ETA~\cite{eata} (a more efficient version of TENT), and SAR~\cite{sar}, which trains the learnable parameters of the batch normalization layers; distillation approaches, such as CoTTA~\cite{cotta}, Pseudo Labeling~(PL)~\cite{pl}, and the very recent and efficient 
LAME~\cite{lame};
EATA~\cite{eata} and % the clustering approach 
TTAC~\cite{ttac} that assume access to the source training data;
data-dependent approaches such as MEMO~\cite{memo} and the diffusion-based method DDA~\cite{dda}.
For all methods, we use their official implementation with their recommended hyper-parameters.
We report our experimental results on a subset of 12 baselines, while leaving ETA, SHOT-IM, and PL to the appendix due to space constraints and their similarity to SHOT and EATA.

As mentioned in Section~\ref{sec:real_time_tta}
, our protocol performs an online computation of the relative adaptation speed of $g$.
In particular, for each batch revealed by the stream, we compute $\mathcal C\left(g(x)\right)$. %  for each received batch by the adaptation method. %  and employ our evaluation protocol in Sec.~\ref{sec:real_time_tta}.
Then, if $\mathcal C(g(x_i)) = k$, all the samples $\{x_{i+1}, x_{i+2}, \dots, x_{i+k}\}$ are processed by $f_{\theta_i}$ without adaptation. 
Otherwise, if $\mathcal C(g(x_i)) = 1$, then these samples are processed by $g$.
For methods that accumulate parameter updates such as TENT~\cite{tent}, $f_{\theta_i}$ is the most recent updated model $g(f_{\theta_{i-1}})$.
% Since several TTA methods' performance depends on the order of the presented data, 
We report all our main results as the average across three seeds, and leave the detailed analysis to the Appendix.
Throughout the experiments, we refer to our realistic evaluation protocol as ``realistic/online", and refer to the current protocol as ``current/offline".
Next, we evaluate all methods on four different scenarios: 
\textit{(i)} when domain shifts happen in an episodic manner, 
% (Section~\ref{sec:episodic_eval}), 
\textit{(ii)} when domain shifts happen continually, \emph{i.e.} one after the other, 
% (Section~\ref{sec:continual_eval}), 
\textit{(iii)} when the stream speed varies, 
% (Section~\ref{sec:stream_speed}), 
\textit{(iii)} when domain shifts happen continually with label correlation; practical evaluation~\cite{rotta}
,and 
\textit{(v)} when the baseline $f_\theta$ is unavailable for evaluating the samples skipped by the TTA method $g$ (left for the appendix).

\begin{figure*}[t]
     \centering
      \begin{subfigure}[b]{1.5\columnwidth}
      % \vspace{-0.1cm}
         \centering
         \includegraphics[width=1.0\columnwidth]{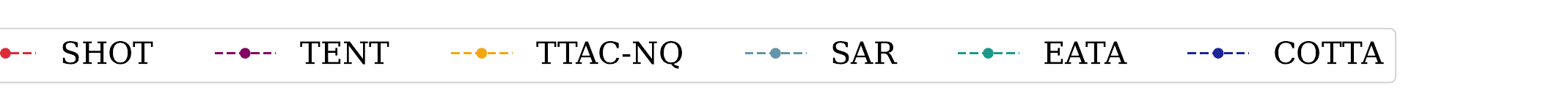}
         % \caption{\textbf{Online} Continual TTA.}
         % \vspace{-0.65cm}
     \end{subfigure}
     % \vspace{-0.2cm}
     \begin{subfigure}[b]{1.0\columnwidth}
         \centering
         \includegraphics[width=1.0\columnwidth]{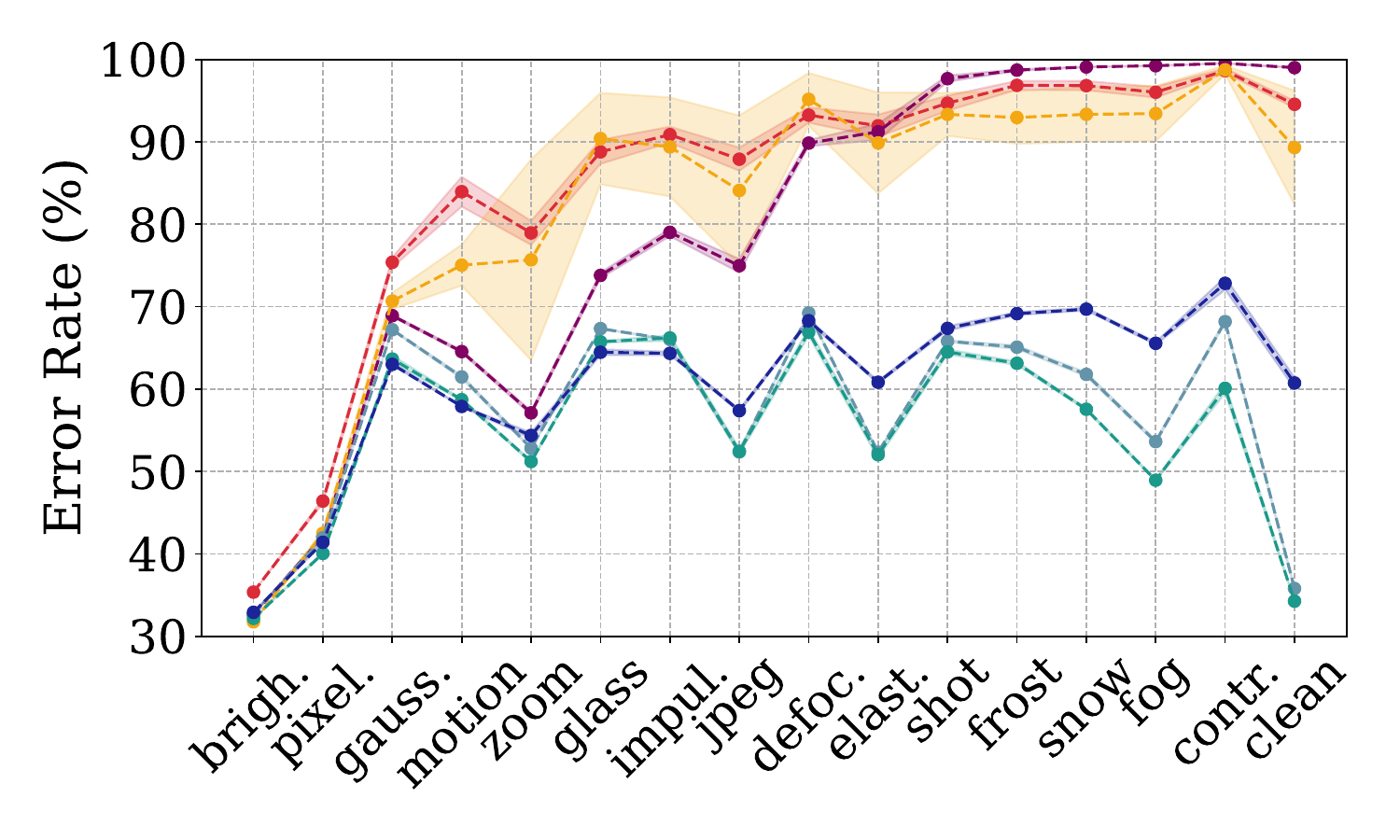}\vspace{-0.2cm}
        \caption{\textbf{Current} Continual TTA.}
     \end{subfigure}
     % \hfill
     \begin{subfigure}[b]{1.0\columnwidth}
         \centering
         \includegraphics[width=1.0\columnwidth]{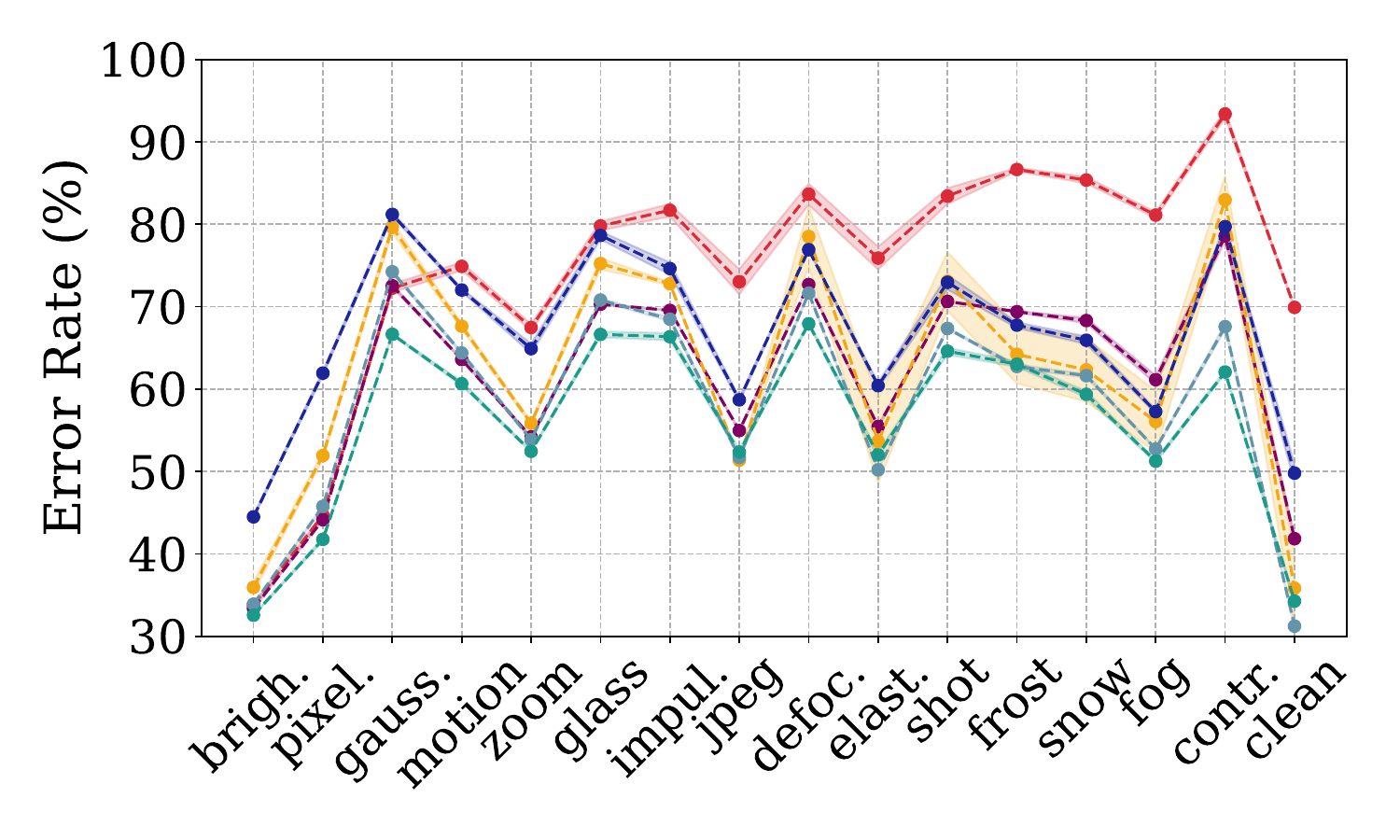}\vspace{-0.2cm}
         \caption{\textbf{Realistic} Continual TTA.}
     \end{subfigure}

     \caption{\textbf{Continual Error Rate on ImageNet-C.} We report the continual error rate of several TTA methods on ImageNet-C benchmark under both realistic and current setups. A lower error rate indicates a better TTA method. Continual evaluation means the corruptions are presented in a sequence without resetting the model in between. We choose the same order as presented along the x-axis; starting with brightness and ending with clean validation set. In the current setup, we observe an increasing trend for SHOT, TENT, and TTAC-NQ. This is hypothesized to be due to overfitting on the early distribution shifts. This behavior is mitigated in the realistic setup due to adapting to fewer batches. EATA and SAR perform equally well in both realistic and current continual setups due to sample rejection. We report the standard deviation across 3 seeds. } \label{fig:continual_setup}
\end{figure*}

\subsection{Episodic Evaluation of TTA}\label{sec:episodic_eval}
First, we consider an episodic evaluation of domain shifts, whereby $\mathcal S$ contains a single domain (\emph{e.g.} one corruption) from ImageNet-C.
We analyze this simple and most common setup to assess the performance of TTA methods under real-time evaluation.
% That is, we want to compare all methods under their proposed original setup, or an easier one.
% For example, methods like MEMO and DDA shine under very small batch sizes, they are also expected to perform well under large batch sizes.
We report the error rates on all corruptions in Table~\ref{tab:imagenetc_episodic} and the average error rate across corruptions.
We summarize the insights as follows:

\noindent\textbf{(i) The performance of TTA methods often degrades significantly under the realistic setup.} 
Most methods induce a significant computational overhead, which prevents them from adapting to every sample from the test stream.
For example, the error rate increases by $7.5\%$ for TTAC-NQ and $4.3\%$ for TENT, where $\mathcal C(g_{\text{TTAC-NQ}}) = 12$ and $\mathcal C(g_{\text{TENT}})=3$ (see Table~\ref{tab:cg_methods}).
That is, TENT adapts to one-third of the batches revealed by the stream, while TTAC-NQ adapts to one every twelve batches.
% Note that in this setup, models that perform cumulative updates (\emph{e.g.} Tent) degrade less than data-dependent approaches.
% This is because $f_{\theta_t}$ for Tent is an updated version while $f_{\theta_t}$ for MEMO is the original source classifier.

\noindent\textbf{(ii) Very efficient methods, with $\mathcal C(g) = 1$, such as LAME and BN, do not lose in performance.}
Evaluating such methods in offline or realistic setups is inconsequential, as their adaptation incurs negligible additional computation (since they adapt during the forward pass~\cite{adabn,bnadaptation} or by adjusting the logits \cite{lame} at a speed that pales in comparison to that of the stream). Interestingly, in our realistic evaluation, the simple BN~(published in 2020) with an average error rate of 66.7\% outperforms more recent and advanced methods such as SAR~(published in 2023) by 1.7\%.
Furthermore, AdaBN~(published in 2017) significantly outperforms the very recent diffusion-based DDA by a notable 13\%.

\noindent\textbf{(iii) Data-dependent approaches, such as MEMO and DDA, are extremely inefficient.}
Despite the independence of MEMO and DDA on batch size, they incur a massive computational burden.
For instance, $\mathcal{C}(g_{\text{MEMO}}) = 54$ and $\mathcal{C}(g_{\text{DDA}}) = 810$.
Thus, both methods will be busy adapting for considerable portions of the stream, leaving most predictions to the non-adapted classifier.
This phenomenon is the reason behind the reported performance of these methods being so close to that of $f_\theta$ (\emph{i.e.} around 82\%). % , which is the same as the source model's performance.
This result calls for future research to focus on increasing the efficiency of data-dependent adaptation methods.

\noindent\textbf{(iv) Sample rejection-oriented methods can perform well under the realistic protocol.}
% Both ETA and 
EATA adapts efficiently due to its fast sample rejection algorithm, which relies solely on the forward pass to admit samples for adaptation. 
EATA's low error rate of 55.6\%, combined with a small performance drop of less than 4\%, positions it as the \textit{top performer} under the realistic evaluation protocol on ImageNet-C. %  dataset under the real-time online evaluation.
% Notably, ETA attains similar performance to EATA, despite the fact that  EATA requires access to the source domain while ETA does not ; % , operates without accessing the source domain, attains similar performance to EATA under this setup and we 
% please refer to the appendix for details on ETA.
On the other hand, SAR does not benefit from sample rejection.
% because of its low performance. 
% also performs sample rejection, it requires the magnitude of the gradient. 
SAR's performance drop of 7.5\% is due to its dependence on gradients for sample rejection, which reduces its speed. % , which results in a performance drop .

\noindent\textbf{(v) SHOT benefits from the realistic protocol.}
Interestingly, we found that SHOT (and SHOT-IM in the Appendix), a fine-tuning-based approach, % the feature extractor, 
benefits from our realistic evaluation.
In particular, we found that SHOT's error rate decreases by 2\% on fog corruption and by 0.8\% on average.
This observation could suggest that SHOT could potentially improve performance by disposing of fine-tuning on every batch. %  without worse in case of adapting to a single distribution shift.
It is also worth mentioning that, under our realistic evaluation, SHOT (introduced in 2020) outperforms \textit{all} methods except EATA.

% \red{Are we missing any other observations? please feel free to drop them.}
\noindent\textbf{(vi) Performance changes are consistent across corruptions.}
Note that all methods that are somewhat efficient can improve the source model across all corruptions, in both the offline and realistic setups. 
Furthermore, the performance changes when comparing the offline and realistic setups are consistent across all corruptions. 
This finding suggests that the performance of these methods is independent of the domain shift being considered. 
We further test this hypothesis by benchmarking these methods on two other  datasets with other types of domain shifts in Section~\ref{sec:other_benchmarks}.

% \begin{enumerate}
%     \item Online evaluation degrades the performance of un-efficient methods.
%     \item Very efficient methods do not get affected by performance degradation.
%     \item SHOT (from early 2020) outperforms Tent, SAR, COTTA, and TTAC, which are all published afterward.
%     \item EATA is indeed an efficient approach. The real-time evaluation did not degrade their performance a lot. This is because their sample rejection algorithm depends on the forward pass computation, unlike SAR which depends on the gradients' magnitude.
%     \item Data-dependent approaches are extremely not efficient. Their online performance is as good as non-adapting. 
%     \item 
% \end{enumerate}

% In the episodic evaluation, the adaptation method is presented with one corruption at a time. When moving to the following corruption, the model is reset back to the base model. We report the error rate for both our reproduction of every method and its corresponding online error rate. As seen in Table \ref{table:episodic}, the best reproduced method is EATA with an average error rate of $52\%$. In online evaluation we can see a significant drop in performance for most methods

% Here we should present the main results that we currently have (online delay vs baseline) for each corruption.
% \shyma{ It is observed that all methods perform the best on the brightness corruption}

\subsection{Continual Evaluation of TTA}
\label{sec:continual_eval}

% Maybe here we will have several setups: \textbf{(i)} continual evaluation where we concat domains. \textbf{(ii)} continual evaluation where clean data is introduced few times. \textbf{(iii)} continual evaluation where everything is mixed. I believe that one of these suffices for the main paper. 
Next, we analyze the more challenging continual setup, following~\cite{cotta, eata}.
In particular, we construct the stream $\mathcal S$ by concatenating \textit{all} corruptions from ImageNet-C.
That is, we adapt TTA methods continually on all corruptions followed by the clean validation set, without ever resetting the network weights. %  at any point.
% Continual evaluation is benchmarked by adapting continuously as the data distribution shifts by each corruption as shown in \cite{cotta,eata,gongnote} \red{CITE MORE PAPERS THAT PRESNETED THE CONTINUAL SETUP}. Where, after each corruption, the model continues to adapt without resetting.
We introduce the notion of realistic adaptation to the continual setup to study the effects of a constant stream speed on the benchmark. 
We report results in Figure~\ref{fig:continual_setup} for both the offline and realistic protocols, where the horizontal-axis shows how corruptions are ordered in the stream. %  presented to the model. 
We limit the experiments in this section to six TTA methods (SHOT, TENT, TTAC-NQ, COTTA, EATA, and SAR), and leave the remaining details for the Appendix. We observe:

\textbf{(i) Methods that do not perform sample rejection (SHOT, TENT, TTAC) scale poorly in the offline-continual setup.}
This phenomenon can be attributed to these methods over-fitting to early distributions.
However, methods that do perform sample rejection (SAR and EATA) do not overfit as easily to corruptions, and can thus adapt to the rest of the stream.
Even worse, such methods tend to even significantly degrade  the performance on clean data.

\textbf{(ii) In the realistic-continual setup, methods that do not perform sample rejection benefit from skipping adaptation on some batches, and become competitive with the methods that perform sample rejection.}
That is, while skipping parts of the stream deteriorated the performance of such methods in the episodic evaluation 
% (in previous Section~\ref{sec:episodic_eval})
, this skipping actually helped in preventing these methods from over-fitting in the continual setup. %  helped these methods to not over-fit to each observed distribution. 

% Thus, adapting to another corruption

% In Figure~\ref{fig:continual_offline} we observe that methods that do not perform sample rejection (SHOT, TENT, TTAC) do not scale well in the offline-continual setup. However, methods that perform sample rejection (SAR, EATA) do not overfit to the current corruption and can then adapt to the rest of the stream. 

% In the online-continual setup presented in Figure~\ref{fig:continual_online}, methods that do not perform sample rejection benefit from skipping adaptation on some batches and become competitive with the methods that perform sample rejection. Interestingly, SHOT does not perform well under the continual setup, despite being one of the best methods during the episodic evaluation. SHOT~\cite{shot} performs updates on the whole feature extractor, giving it more parameters to overfit to the different corruptions and making it more challenging to adapt for the remaining.

% present in table \ref{lifelongimgntc} and table \ref{lifelongimgnt3dcc}.

\subsection{Stream Speed Analysis}
\label{sec:stream_speed}

\begin{figure}[t]
    \centering
    \includegraphics[width=1.0\columnwidth]{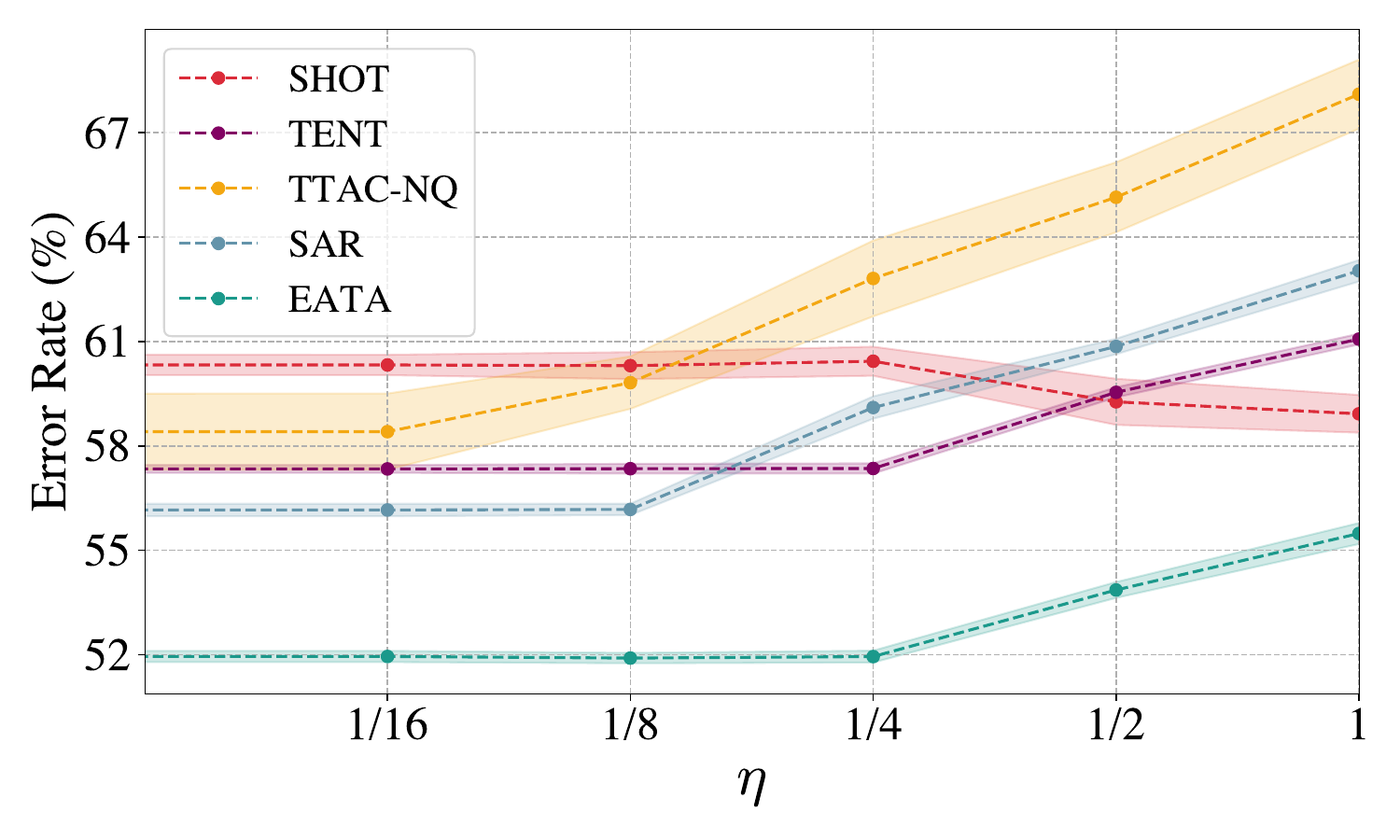}\vspace{-0.15cm}
    \caption{\textbf{Average Error Rate on ImageNet-C Under Slower Stream Speeds.} We report the average error rate for several TTA methods on ImageNet-C under slower stream speeds. In our proposed realistic model evaluation, the stream speed $r$ is normalized by the time needed for a forward pass using the base model. 
    % However, this might not be true, depending on the application. 
    % The stream speed can actually be slower. 
    We evaluate different TTA methods under a stream with speed $\eta r$ with $\eta \in (0, 1]$. 
    % We define a ratio $\eta$ that controls the speed stream. 
    An $\eta=\nicefrac{1}{16}$ means the stream is $16$ times slower than the forward pass of the base model. 
    % If the stream is faster, \ie $\eta>1$, we argue that the reasonable course of action is to change the base model to a faster one. 
    We report the standard deviation across 3 different random seeds.
    Different TTA methods degrade differently when varying $\eta$.
    % \juan{the $\eta$ in the xlabel looks too bold}
    }
    \vspace{-0.25cm}
    \label{fig:stream_speed_standard}
\end{figure}
In the previous experiments, we normalized the stream speed to be the same as that of $f_\theta$'s forward pass. 
That is, we assumed % in Section~\ref{sec:real_time_tta} 
that the rate $r$ at which $\mathcal S$ reveals new batches is equal to $R\left(f_\theta(x)\right)$.
%\juan{should we mention something here like "which is assuming that the stream is very fast"? I think this would then allow us to say that at $\eta = 1$ we recover our protocol, where the stream is "fast". However, this claim would also call for us to cite something stating that the forwards of current models are actually fast enough to be deployed in practice (which I'm 99\% sure it is 100\% true)}.
However, % this assumption might not hold for 
some applications may enjoy a slower stream, %  might reveal data at a slower rate, 
giving TTA methods more time to adapt to samples. % the current batch.
To explore this scenario, we vary the speed at which the stream reveals new data.
In particular, let the new stream rate be $\eta\:r$ % =\eta R\left(f_\theta\right)$ 
with $\eta \in (0, 1]$.
Hence, as $\eta \rightarrow 0$, the stream slows down and allows methods to adapt to all samples.
Conversely, as $\eta \rightarrow 1$, the stream speeds up, and at $\eta = 1$ we recover our realistic evaluation protocol. %  where the stream is very fast.

We experiment with the stream speed by setting $\eta\in\{\nicefrac{1}{16}, \nicefrac{1}{8}, \nicefrac{1}{4}, \nicefrac{1}{2}, 1\}$, and evaluate five representative TTA methods (SHOT, TENT, TTAC-NQ, SAR, and EATA) in the episodic setup .
% (similar to Section~\ref{sec:episodic_eval}). %  under different stream speeds by .
Figure~\ref{fig:stream_speed_standard} summarizes our results by reporting the average error rate across all corruptions.
% and report the average error rate across different corruptions in Figure~\ref{fig:stream_speed_single}. 
We next list our observations:

\textbf{(i) The performance of TTA methods varies widely.}
For example, TTAC-NQ starts degrading faster (at $\eta=\nicefrac{1}{16}$) due to its slow adaptation speed. % higher adaptation complexity.
For other methods, the $\eta$ at which they degrade varies.
 For instance, while TENT has a higher error rate than SAR in slow streams ($\eta \leq \nicefrac{1}{8}$), TENT outperforms SAR in the regime of faster streams $\eta \leq \nicefrac{1}{4}$.
 Interestingly, SHOT~\cite{shot} ranks the worst at $\eta \leq \nicefrac{1}{8}$, then ranks second when $\eta \geq \nicefrac{1}{2}$, becoming a viable alternative.
 At last, the order of different methods significantly changes depending on the speed of the stream.
 For example, SAR changes from being second best at $\eta\leq\nicefrac{1}{8}$ to third at $\eta=\nicefrac{1}{4}$ and then to fifth (\emph{i.e.} second worst) at $\eta \geq \nicefrac{1}{2}$.
% That is, while the offline ranking of the considered methods stays the same for $r \leq 1/8$, 
 
\textbf{(ii) EATA provides a good trade-off between speed and performance.} % \juan{I think I understand what was meant here, but the statement sounds kind of inconsistent: speed and what we called ``adaptation complexity'' are inversely related with a one-to-one mapping, no? There can be no ``good trade-off'', ad the trade-off is fixed.}} 
In fact, EATA gives the best overall performance (lowest error rate) independent of the stream's speed.
This virtue is attributable to EATA's combination of good performance and adaptation speed based on efficient sample rejection.
Results on other datasets are in the Appendix. %  which we leave for the appendix.

\subsection{Results on Other Benchmarks and Architectures}
\label{sec:other_benchmarks}
% \usepackage{multirow}
% \usepackage[table.xcdraw]{xcolor}
% If you use a beamer only pass "xcolor=table" option. i.e. \documentclass[xcolor=table]{beamer}
% \captionsetup{font={small,it}}
\begin{table*}[t]
\caption{\textbf{Episodic Error Rate on ImageNet-C with ViT.}
We report the error rate of three baselines (Source, Tent, SAR) on the 15 different corruptions on ImageNet-C when the backbone is ViT architecture pretrained on ImageNet. We observe that while generally better backbones yield smaller error rate, expensive methods perform worse under our realistic evaluation. The more expensive the method is (e.g. SAR compared to Tent), the more performance reduction it suffers.
}
% \vspace{-0.1cm}
\label{tab:imagenetc_episodic_vit} 
\resizebox{\linewidth}{!}{
% \usepackage{multirow}
% \usepackage[table.xcdraw]{xcolor}
% If you use beamer only pass "xcolor=table" option. i.e. \documentclass[xcolor=table]{beamer}
\begin{tabular}{l|c|ccc|cccc|cccc|cccc|c|c}
\toprule
\rowcolor[HTML]{FFFFFF} 
\cellcolor[HTML]{FFFFFF}                          & \cellcolor[HTML]{FFFFFF}                        & \multicolumn{3}{c|}{\cellcolor[HTML]{FFFFFF}Noise} & \multicolumn{4}{c|}{\cellcolor[HTML]{FFFFFF}Blur}                           & \multicolumn{4}{c|}{\cellcolor[HTML]{FFFFFF}Weather} & \multicolumn{4}{c|}{\cellcolor[HTML]{FFFFFF}Digital} & \cellcolor[HTML]{FFFFFF}                       \\
\rowcolor[HTML]{FFFFFF} 
\multirow{-2}{*}{\cellcolor[HTML]{FFFFFF}Method}  & \multirow{-2}{*}{\cellcolor[HTML]{FFFFFF}Realistic} & gauss.                         & shot   & impul.  & defoc.                       & glass & motion                       & zoom & snow       & frost       & fog        & brigh.      & contr.      & elast.      & pixel.      & jpeg      & \multirow{-2}{*}{\cellcolor[HTML]{FFFFFF}Avg.} & \multirow{-2}{*}{\cellcolor[HTML]{FFFFFF}$\Delta$}\\
\midrule
\rowcolor[HTML]{FFFFFF} 
Source                                            & \cmark                                          &90.5 &        93.3 &           91.8 &          71.0 &        76.6 &         66.1 &       72.9 &  84.1 &   73.5 &  52.8 &        45.3 &      55.9 &               69.5 &      55.5 &              52.2 &     70.1  & -                                           \\%\midrule \midrule
\midrule
% \rowcolor[HTML]{FFFFFF}
\rowcolor[HTML]{e9f5f9} 
\cellcolor[HTML]{FFFFFF}                          & \cellcolor[HTML]{FFFFFF}\xmark                 & 69.9 &      \cellcolor[HTML]{FFFFFF}  95.9 &           68.9 &         55.8 &      62.0 &        \cellcolor[HTML]{FFFFFF} 52.3 &  \cellcolor[HTML]{FFFFFF}     57.9 & \cellcolor[HTML]{e9f5f9}  57.2 &   53.6 &  41.8 &        28.9 &      40.7 &               59.1 &      39.7 &              42.0 &     55.0 & \cellcolor[HTML]{FFFFFF}                                            \\
% \rowcolor[HTML]{e9f5f9} 
\multirow{-2}{*}{Tent}    & \cmark                  &80.7 &      \cellcolor[HTML]{e9f5f9}   88.9 &    81.0 &      63.0 &       69.5 &      \cellcolor[HTML]{e9f5f9}    58.3 & \cellcolor[HTML]{e9f5f9}      64.9 &  65.8 &59.7 &  47.7 &        33.2 &      47.3 &               64.6 &      45.1 &              46.4 &     61.1 & \multirow{-2}{*}{\cellcolor[HTML]{FFFFFF} \textcolor{red}{(-6.1)}}                                  \\
\midrule
\rowcolor[HTML]{e9f5f9} 
\cellcolor[HTML]{FFFFFF}                          & \cellcolor[HTML]{FFFFFF}\xmark                 & 55.5 &        56.9 &           55.1 &          47.5 &        50.4 &         44.3 &       48.7 &  42.4 &   47.3 &  33.6 &        25.4 &      35.6 &               44.8 &      33.5 &              36.4 &     43.8&        \cellcolor[HTML]{FFFFFF}                                  \\
\rowcolor[HTML]{FFFFFF} 
\multirow{-2}{*}{\cellcolor[HTML]{FFFFFF}SAR}    & \cellcolor[HTML]{FFFFFF}\cmark                  &70.0 &        72.5 &           69.4 &          56.6 &        63.4 &         54.0 &       60.0 &  56.4 &   53.5 &  43.0 &        30.5 &      43.3 &               58.7 &      41.5 &              43.8 &     54.5 & \multirow{-2}{*}{\textcolor{red}{(-10.7)}}                                          \\                                           
                                      \midrule \bottomrule
\end{tabular}
}
% \vspace{-10pt}
\end{table*}

% \captionsetup{font={small,it}}

% To evaluate consistency across other benchmarks rather than being specific to ImageNet-C. 

We extend our evaluation protocol to cover ImageNet-3DCC~\cite{3dcc} and ImageNet-R~\cite{imagenetr} datasets and ResNet-18 (results in the appendix) and ViT~\cite{vit} architectures.
% We use ImageNet-3DCC~\cite{3dcc} and ImageNet-R~\cite{imagenetr}. \red{add more details about what ImageNet-3DCC and ImageNet-R are} 
ImageNet-R contains rendition versions of ImageNet spanning 200 classes.
ImageNet-3DCC constructs more spatially-consistent corruptions than ImageNet-C by leveraging depth estimates. % and 3D priors.
For ViT, we conduct episodic evaluation on ImageNet-C in a similar setup to Section~\ref{sec:episodic_eval} and report the results in Table~\ref{tab:imagenetc_episodic_vit} for the non-adapted model, Tent, and SAR.
For ImageNet-R and ImageNet-3DCC, we fix the architecture to ResNet-50 and experiment on the entire datasets and set the severity level to 5 in ImageNet-3DCC.
Due to the space constraint, we limit our experiments to the episodic evaluation, 
% similar to Section~\ref{sec:episodic_eval}, 
and leave other results and analyses to the Appendix.
We evaluate the effectiveness of 10 TTA methods in Table~\ref{tab:imagenetr_and_3dcc}, where we report the average error rate across all corruptions.
% We observe:
% We use the original offline setup \ref{sec:test time adaptation} and compare it with our online setup \ref{sec:real_time_tta}. We report both online and offline average across all corruptions in Table~\ref{tab:imagenetr_and_3dcc} for the methods reported in ImagenetC. 

We observe that \textbf{our results are consistent across all considered datasets and architectures.}
Similar to our results in Table~\ref{tab:imagenetc_episodic}, the more computationally involved SAR degrades more than Tent when leveraging ViT architecture.
% and Figure~\ref{fig:stream_speed_standard}. 
Regarding other datasets, we find that simple methods that adapt during the forward pass are unaffected by the realistic setup. 
All the other methods, except SHOT, experience degradation in their results on both datasets. 
We observe again that, on these two datasets, while SHOT actually benefits from the realistic evaluation,
% and becomes the top performer on ImageNet-3DCC by surpassing both EATA and SAR. 
% On ImageNet-R, 
EATA remains the best alternative on both ImageNet-R and ImageNet-3DCC.

% \usepackage{multirow}
\begin{table}[t]
\caption{\textbf{Average Error Rate on ImageNet-R and ImageNet-3DCC.}
We report the average error rate of different TTA methods on ImageNet-R and ImageNet-3DCC under both the realistic and current setups. A lower error rate indicates a better TTA method. The highlighted numbers indicate a better performance per method across setups. We observe that methods generally perform worse in the more realistic realistic setup. The conclusions are consistent with what we observed on ImageNet-C~(Table~\ref{tab:imagenetc_episodic}).
% \red{We need to be consistent with the subset of methods we're reporting on. We need to be consisted with the methods grouping} 
}
% \vspace{-0.15cm}
\footnotesize
\centering
{\setlength{\tabcolsep}{3.0pt}
\resizebox{\columnwidth}{!}{
\begin{tabular}{l|ccc|ccc}
\toprule
\multicolumn{1}{l|}{\multirow{2}{*}{Method}} & \multicolumn{3}{c|}{\textbf{ImageNet-R}} & \multicolumn{3}{c}{\textbf{ImageNet-3DCC}} \\
\multicolumn{1}{c|}{}                        & Current       & Realistic  & $\Delta$      & Current         & Realistic   & $\Delta$      \\
\midrule Source                                       &  63.8            &  63.8  & -       & 73.9             &  73.9  &  -         \\

AdaBN                                      &  60.6            & 60.6  & 0       & {72.1}             & {72.1}    & 0       \\
BN                               &  60.0            & 60.0  & 0       & 70.5             & 70.5     & 0      \\
LAME                                        & 60.5            & 60.5  & 0       & 72.1             & 72.1    & 0       \\
% PL                                          & 56.2            & 57.5         & 70.3             & 68.7           \\
\midrule 
SHOT                                       & 70.3            & \cellcolor[HTML]{e9f5f9} 62.6     &   \textcolor{GTgreen}{(+7.7)}  & 69.2          & \cellcolor[HTML]{e9f5f9}67.0  & \textcolor{GTgreen}{(+2.2)}         \\
% SHOT-IM                                     & 69.4            & 61.8         & 68.5             & 66.6           \\
TENT                                        & \cellcolor[HTML]{e9f5f9}58.1            & 59.1   & \textcolor{red}{(-1.0)}      & \cellcolor[HTML]{e9f5f9}{64.5}             & {66.8}  &   \textcolor{red}{(-2.3)}       \\

SAR                                        & \cellcolor[HTML]{e9f5f9}57.5            & 59.6   & \textcolor{red}{(-2.1)}      & \cellcolor[HTML]{e9f5f9}63.5             & 71.4  & \textcolor{red}{(-7.9)}    \\
CoTTA                                       & \cellcolor[HTML]{e9f5f9}57.3            & 61.5  & \textcolor{red}{(-4.5)}       & \cellcolor[HTML]{e9f5f9}{66.4}             & {75.6}    & \textcolor{red}{(-9.2)}       \\
\midrule 
% ETA                                         & 56.2            & 57.2         & 67.8             & 70.1           \\
EATA                                     & \cellcolor[HTML]{e9f5f9}55.7            & 57.1    & \textcolor{red}{(-1.4)}     & \cellcolor[HTML]{e9f5f9}{60.9}             & {63.1}     & \textcolor{red}{(-2.2)}      \\
TTAC-NQ                                    & \cellcolor[HTML]{e9f5f9}59.2            & 60.8    & \textcolor{red}{(-1.6)}     & \cellcolor[HTML]{e9f5f9}{65.7}             & {73.6}    & \textcolor{red}{(-7.9)}       \\

\bottomrule     
\end{tabular}}
}
% \vspace{-10pt}
\label{tab:imagenetr_and_3dcc}
\end{table}

\subsection{Evaluation under Practical TTA}
Recently, \cite{rotta} extended the continual test-time adaptation evaluation to include label-imbalances; known as Practical Test-Time Adaptation~(PTTA) setup.
In this setting, the stream not only reveals a continual sequence of distribution shifts, but also the revealed batches have significant label imbalances.
To combat this combined challenge, the work of~\cite{rotta} proposed to leverage a balanced memory bank for adaptation.
In this section, we extend our computational constrained evaluation to the PTTA setup and compare RoTTA~\cite{rotta} with a non-adapted model on CIFAR10-C benchmark.

% \usepackage{multirow}
% \usepackage[table.xcdraw]{xcolor}
% If you use a beamer only pass "xcolor=table" option. i.e. \documentclass[xcolor=table]{beamer}
% \captionsetup{font={small,it}}
\begin{table*}[t]
\caption{\textbf{Episodic Error Rate on CIFAR10-C under Practical Evaluation~\cite{rotta}.}
We report the error rate of two baselines (Source, RoTTA~\cite{rotta}) on the 15 different corruptions on CIFAR10-C when the backbone is ResNet-18. 
We observe that under our computational constrained evaluation, the only method tailored to this setting; RoTTA, performs worse than the non-adapted baseline.
}
% \vspace{-0.1cm}
\label{tab:rotta_cifar10c} 
\resizebox{\linewidth}{!}{
% \usepackage{multirow}
% \usepackage[table.xcdraw]{xcolor}
% If you use beamer only pass "xcolor=table" option. i.e. \documentclass[xcolor=table]{beamer}
\begin{tabular}{l|c|ccc|cccc|cccc|cccc|c|c}
\toprule
\rowcolor[HTML]{FFFFFF} 
\cellcolor[HTML]{FFFFFF}                          & \cellcolor[HTML]{FFFFFF}                        & \multicolumn{3}{c|}{\cellcolor[HTML]{FFFFFF}Noise} & \multicolumn{4}{c|}{\cellcolor[HTML]{FFFFFF}Blur}                           & \multicolumn{4}{c|}{\cellcolor[HTML]{FFFFFF}Weather} & \multicolumn{4}{c|}{\cellcolor[HTML]{FFFFFF}Digital} & \cellcolor[HTML]{FFFFFF}                       \\
\rowcolor[HTML]{FFFFFF} 
\multirow{-2}{*}{\cellcolor[HTML]{FFFFFF}Method}  & \multirow{-2}{*}{\cellcolor[HTML]{FFFFFF}Realistic} & gauss.                         & shot   & impul.  & defoc.                       & glass & motion                       & zoom & snow       & frost       & fog        & brigh.      & contr.      & elast.      & pixel.      & jpeg      & \multirow{-2}{*}{\cellcolor[HTML]{FFFFFF}Avg.} & \multirow{-2}{*}{\cellcolor[HTML]{FFFFFF}$\Delta$}\\
\midrule
\rowcolor[HTML]{FFFFFF} 
Source                                            & \cmark                                          &72.3	&65.7	&72.9	&46.9	&54.3	&34.8	&42.0	&25.1	&41.3	&26.0	&9.3	&46.7	&26.6	&58.5	&30.3	&43.5 & -                                           \\%\midrule \midrule
\midrule
% \rowcolor[HTML]{FFFFFF}
\midrule
\rowcolor[HTML]{e9f5f9} 
\cellcolor[HTML]{FFFFFF}                          & \cellcolor[HTML]{FFFFFF}\xmark                 & 36.9	&34.9	&45.8	&16.6	&44.2	&19.9	&16.53	&21.6	&22.4	&18.8	&9.8	&20.6	&28.4	&27.1	&34.5&	26.5	&        \cellcolor[HTML]{FFFFFF}                                  \\
\rowcolor[HTML]{FFFFFF} 
\multirow{-2}{*}{\cellcolor[HTML]{FFFFFF}RoTTA}    & \cellcolor[HTML]{FFFFFF}\cmark                  &55.0	 &54.4	 &63.2	 &43.3	 &62.3	 &43.7	 &43.5	 &44.8	 &47.7	 &43.4	 &35.3	 &41.8	 &54.0	 &47.7	 &54.6	 &49.0 & \multirow{-2}{*}{\textcolor{red}{(-22.5)}}                                          \\                                           
                                      \midrule \bottomrule
\end{tabular}
}
% \vspace{-10pt}
\end{table*}

% \captionsetup{font={small,it}}

Table~\ref{tab:rotta_cifar10c} summarizes the results.
We observe that while RoTTA indeed reduces the error rate under the PTTA setup on CIFAR10-C (17\% below the non-adapted model), our realistic evaluation uncovers its computational limitation.
We found that RoTTA's error rate increases by over 22\% surpassing the error rate of the non-adapted model.
Note that RoTTA stores samples from the stream in a memory bank then adapts the model on sampled samples from the memory bank. Thus, the slower the adaptation of RoTTA, the less diverse the samples in the memory bank, hindering its adaptation.

% \subsection{Analysis}
% In this section, we present a detailed analysis of our evaluation.
% We begin by assessing the impact of hyper-parameter tunning on the performance drop under our proposed evaluation.
% Then, we provide a discussion on how do different TTA methods spend their compute budget during the inference.
% We conclude with some potential recommendations for future works on developing TTA methods that are both accurate and efficient.

\subsection{Effect of Hyper-parameter Tuning}
The performance of different TTA methods heavily depends on their hyper-parameter settings~\cite{zhao2023pitfalls}.
Here, we assess the impact of our proposed evaluation on TTA methods when tuning their hyperparameters.
For that regard, we conduct hyper parameter search for Tent (as a fundamental baseline) and experiment with different learning rates (the only hyper-parameter for Tent).

Table~\ref{tab:hyperparameters_tent} summarizes the results under episodic evaluation for 4 different corruptions on ImageNet-C.
We observe that while conducting hyper-parameter search indeed improves the performance of TENT, its error rate increases under our realistic evaluation across all hyperparameters. That is, while conducting hyper-parameter search might indeed result in a better performance for TTA methods, the insights obtained through our proposed evaluation scheme remains consistent: more efficient TTA methods will have a smaller performance drop under the realistic evaluation. 

% \subsubsection{Where Do TTA Methods Spend Their Compute?}
% \usepackage{multirow}
% \usepackage[table.xcdraw]{xcolor}
% If you use a beamer only pass "xcolor=table" option. i.e. \documentclass[xcolor=table]{beamer}
% \captionsetup{font={small,it}}
\begin{table}[t]
\caption{\textbf{Effect of our evaluation under hyperparameter tuning.}
We report the error rate for Tent under different learning rates under both the current and our proposed realistic evaluation.
While carefully tuning the learning rate for Tent results in a better performance, our realistic evaluation causes a performance drop under all  learning rates.
}
% \vspace{-0.1cm}
\label{tab:hyperparameters_tent} 
\resizebox{\linewidth}{!}{
% \usepackage{multirow}
% \usepackage[table.xcdraw]{xcolor}
% If you use beamer only pass "xcolor=table" option. i.e. \documentclass[xcolor=table]{beamer}
\begin{tabular}{l|c|cccc|c|c}
\toprule
\rowcolor[HTML]{FFFFFF} 
{\cellcolor[HTML]{FFFFFF}$lr$}  & {\cellcolor[HTML]{FFFFFF}Realistic} & gauss.                         &motion                       &  fog        & pixel.         & {\cellcolor[HTML]{FFFFFF}Avg.} & {\cellcolor[HTML]{FFFFFF}$\Delta$}\\
\midrule
% \rowcolor[HTML]{FFFFFF}
\midrule
\rowcolor[HTML]{e9f5f9} 
\cellcolor[HTML]{FFFFFF}                          & \cellcolor[HTML]{FFFFFF}\xmark                &74.1	&63.3	&44.7	&43.5&	56.4	&        \cellcolor[HTML]{FFFFFF}                                  \\
\rowcolor[HTML]{FFFFFF} 
\multirow{-2}{*}{\cellcolor[HTML]{FFFFFF} $1\times 10^{-4}$}    & \cellcolor[HTML]{FFFFFF}\cmark                  &	79.7	&69.0	&47.8	&46.8	&60.8	 & \multirow{-2}{*}{\textcolor{red}{(-4.4)}}                                          \\                                           
                                      \midrule 
                                      \rowcolor[HTML]{e9f5f9} 
\cellcolor[HTML]{FFFFFF}                          & \cellcolor[HTML]{FFFFFF}\xmark                &71.1	&59.7&	43.1&	41.9&	53.9	&        \cellcolor[HTML]{FFFFFF}                                  \\
\rowcolor[HTML]{FFFFFF} 
\multirow{-2}{*}{\cellcolor[HTML]{FFFFFF} $2\times 10^{-4}$}    & \cellcolor[HTML]{FFFFFF}\cmark                  &		77.6&	66.1	&46.0	&45.0&	58.7	 & \multirow{-2}{*}{\textcolor{red}{(-4.7)}}                                          \\                                           
                                      \midrule 
                                      \rowcolor[HTML]{e9f5f9} 
\cellcolor[HTML]{FFFFFF}                          & \cellcolor[HTML]{FFFFFF}\xmark                &69.6	&58.1	&42.4&	41.1	&52.8	&        \cellcolor[HTML]{FFFFFF}                                  \\
\rowcolor[HTML]{FFFFFF} 
\multirow{-2}{*}{\cellcolor[HTML]{FFFFFF} $3\times 10^{-4}$}    & \cellcolor[HTML]{FFFFFF}\cmark                  &	74.9	&64.0	&45.0	&44.0	&57.0	 & \multirow{-2}{*}{\textcolor{red}{(-4.2)}}                                          \\                                           
                                      \midrule 
                                      \rowcolor[HTML]{e9f5f9} 
\cellcolor[HTML]{FFFFFF}                          & \cellcolor[HTML]{FFFFFF}\xmark                &	68.8	&57.1	&42.0	&40.8	&52.2	&        \cellcolor[HTML]{FFFFFF}                                  \\
\rowcolor[HTML]{FFFFFF} 
\multirow{-2}{*}{\cellcolor[HTML]{FFFFFF} $4\times 10^{-4}$}    & \cellcolor[HTML]{FFFFFF}\cmark                  &	73.7	&62.3	&44.5	&43.2	&55.9	 & \multirow{-2}{*}{\textcolor{red}{(-3.7)}}                                          \\                                           
                                      \midrule 
                                      \bottomrule
\end{tabular}
}
\vspace{-5pt}
\end{table}

\section{Conclusions}
In this work, we find that the performance of Test Time Adaptation (TTA) methods can vary depending on the context in which they are used. 
In the episodic evaluation, the efficiency of the method is the most important factor, with more efficient methods like AdaBN and BN showing consistent performance, while data-dependent approaches suffer. Sample rejection methods generally perform well, and fine-tuning approaches such as SHOT can even improve when adapting to fewer samples.
In the continual evaluation, methods that do not perform sample rejection scale poorly in the offline-continual setup but benefit from skipping adaptation on some batches in the realistic-continual setup. 
Furthermore, our stream speed analysis shows that the performance of TTA methods can vary widely at different speeds.
% but EATA provides a good trade-off between speed and performance.
Our findings are consistent across corruptions and multiple datasets. They can help researchers and practitioners to better understand the strengths and weaknesses of different TTA methods, and to choose the most appropriate method for their specific use case. 
% We hope that our realistic evaluation scheme inspires the development of TTA methods that are both effective and efficient in real-world settings.
\section*{Acknowledgements}
This work was partially done during a research internship of the first author at Intel Labs.
This work was supported by the King Abdullah University of Science and Technology (KAUST) Office of Sponsored Research (OSR) under Award No. OSR-CRG2021-4648. 
We would like to thank Yasir Ghunaim and Mattia Soldan for the helpful discussion.
% \newpage
\section*{Impact Statement}
Our work advances Machine Learning by proposing a realistic evaluation protocol for Test Time Adaptation methods, prioritizing computational efficiency. 
This approach promotes the development of AI systems that are both accessible in resource-limited settings and environmentally sustainable, by favoring simpler, faster methods. 
Such advancements contribute to more inclusive and responsible AI deployment, aligning with ethical goals of broadening access and reducing environmental impacts

\bibliography{example_paper}
\bibliographystyle{icml2024}

%%%%%%%%%%%%%%%%%%%%%%%%%%%%%%%%%%%%%%%%%%%%%%%%%%%%%%%%%%%%%%%%%%%%%%%%%%%%%%%
%%%%%%%%%%%%%%%%%%%%%%%%%%%%%%%%%%%%%%%%%%%%%%%%%%%%%%%%%%%%%%%%%%%%%%%%%%%%%%%
% APPENDIX
%%%%%%%%%%%%%%%%%%%%%%%%%%%%%%%%%%%%%%%%%%%%%%%%%%%%%%%%%%%%%%%%%%%%%%%%%%%%%%%
%%%%%%%%%%%%%%%%%%%%%%%%%%%%%%%%%%%%%%%%%%%%%%%%%%%%%%%%%%%%%%%%%%%%%%%%%%%%%%%
\newpage
\appendix
% \onecolumn
\appendix
% \usepackage{multirow}
% \usepackage[table.xcdraw]{xcolor}
% If you use a beamer only pass "xcolor=table" option. i.e. \documentclass[xcolor=table]{beamer}
\begin{table*}[bp]
\caption{\label{tab:app_episodic} \textbf{Episodic Error Rate on ImageNet-C.}
We report the error rate of different TTA methods on the ImageNet-C benchmark under both the online and offline setups. 
A lower error rate indicates a better TTA method. 
The highlighted numbers indicate a better performance per method across setups. 
Episodic means the model will adapt to one corruption at a time. 
The model is reset back to the base model when moving to the next corruption. 
The offline setup is merely the reproduction of every method. 
We show that methods generally perform worse in the more realistic online setup. 
The more computationally complex the TTA method is, the less data it will adapt to, and the worse its performance. 
SAR-GN represents SAR when deployed on ResNet50 with Group Normalization (GN) layers, following~\cite{sar}.
% \red{we need to devise a better grouping like the one we did in the related work, TENT/SAR/EATA are entropy minimization methods, MEMO/CoTTA use consistency over augmentations, AdaBN/BN adapt batch norm statistics etc.}
}
\resizebox{\textwidth}{!}{
\begin{tabular}{l|c|ccc|cccc|cccc|cccc|c|c}
\toprule
\rowcolor[HTML]{FFFFFF} 
\cellcolor[HTML]{FFFFFF}                          & \cellcolor[HTML]{FFFFFF}                        & \multicolumn{3}{c|}{\cellcolor[HTML]{FFFFFF}Noise} & \multicolumn{4}{c|}{\cellcolor[HTML]{FFFFFF}Blur}                           & \multicolumn{4}{c|}{\cellcolor[HTML]{FFFFFF}Weather} & \multicolumn{4}{c|}{\cellcolor[HTML]{FFFFFF}Digital} & \cellcolor[HTML]{FFFFFF}                       \\
\rowcolor[HTML]{FFFFFF} 
\multirow{-2}{*}{\cellcolor[HTML]{FFFFFF}Method}  & \multirow{-2}{*}{\cellcolor[HTML]{FFFFFF}Online} & gauss.                         & shot   & impul.  & defoc.                       & glass & motion                       & zoom & snow       & frost       & fog        & brigh.      & contr.      & elast.      & pixel.      & jpeg      & \multirow{-2}{*}{\cellcolor[HTML]{FFFFFF}Avg.} & \multirow{-2}{*}{\cellcolor[HTML]{FFFFFF}$\Delta$}\\
\midrule
\rowcolor[HTML]{FFFFFF} 
\cellcolor[HTML]{FFFFFF}                          & \cellcolor[HTML]{FFFFFF}\xmark                 & 73.1	&69.8	&72.0	&76.9	&75.9	&\cellcolor[HTML]{e9f5f9}58.5	&52.7	&53.3	&62.2	&43.8	&34.6	&82.6	&46.0	&42.3	&48.9	&59.5                                            \\
\rowcolor[HTML]{e9f5f9} 
\multirow{-2}{*}{\cellcolor[HTML]{FFFFFF}SHOT-IM}    & \cellcolor[HTML]{FFFFFF}\cmark                  & 71.1&68.6&70.7&73.2&73.6&\cellcolor[HTML]{FFFFFF}59.1&51.9&52.8&60.5&43.7&33.6&77.3&45.7&42.1&48.6      & 58.2 & \multirow{-2}{*}{\cellcolor[HTML]{FFFFFF} \textcolor{green}{(-0.3)}}                                  \\
\midrule
\rowcolor[HTML]{e9f5f9} 
\cellcolor[HTML]{FFFFFF}                          & \cellcolor[HTML]{FFFFFF}\xmark                 & \cellcolor[HTML]{FFFFFF}92.2	&\cellcolor[HTML]{FFFFFF}92.2	&\cellcolor[HTML]{FFFFFF}92.8	&\cellcolor[HTML]{FFFFFF}97.0	&\cellcolor[HTML]{FFFFFF}89.8	&57.7	&49.6	&50.7	&57.1	&41.5	&32.6	&91.1	&44.3	&40.3	&46.6	&65.0                                           \\
\rowcolor[HTML]{FFFFFF} 
\multirow{-2}{*}{\cellcolor[HTML]{FFFFFF}PL}    & \cellcolor[HTML]{FFFFFF}\cmark                  & \cellcolor[HTML]{e9f5f9}90.6	&\cellcolor[HTML]{e9f5f9}86.3	&\cellcolor[HTML]{e9f5f9}83.6	&\cellcolor[HTML]{e9f5f9}93.2	&\cellcolor[HTML]{e9f5f9}89.7	&63.0	&51.7	&55.0	&59.3	&43.8	&32.9	&92.3	&47.3	&42.4	&49.3	&65.3 & \multirow{-2}{*}{\cellcolor[HTML]{FFFFFF} \textcolor{red}{(+0.3)}}                                  \\
\midrule
\rowcolor[HTML]{e9f5f9} 
\cellcolor[HTML]{FFFFFF}                          & \cellcolor[HTML]{FFFFFF}\xmark                 & 64.9	&62.7	&63.6	&66.4	&66.3	&52.4	&47.3	&48.2	&54.1	&40.2	&32.2	&54.8	&42.3	&39.2	&44.7	&52.0                                          \\
\rowcolor[HTML]{FFFFFF} 
\multirow{-2}{*}{\cellcolor[HTML]{FFFFFF}ETA}    & \cellcolor[HTML]{FFFFFF}\cmark                  &70.2	&67.0	&69.6	&71.5	&71.5	&56.9	&50.2	&51.9	&57.0	&42.0	&32.5	&60.5	&44.6	&40.8	&47.1	&55.6  &  \multirow{-2}{*}{\cellcolor[HTML]{FFFFFF}\textcolor{red}{(+3.6)}}                                       \\
\midrule
\rowcolor[HTML]{e9f5f9} 
\cellcolor[HTML]{FFFFFF}                          & \cellcolor[HTML]{FFFFFF}\xmark                 & 71.8	&69.0	&70.3	&81.5	&81.0	&69.6	&69.5	&57.1	&56.6	&94.3	&29.2	&56.0	&84.8	&51.4	&44.7	&65.8                                         \\
\rowcolor[HTML]{FFFFFF} 
\multirow{-2}{*}{\cellcolor[HTML]{FFFFFF}SAR-GN}     & \cellcolor[HTML]{FFFFFF}\cmark                  & 82.0	&80.2	&82.1	&80.2	&88.6	&78.5	&75.1	&59.6	&53.9	&66.9	&30.7	&63.3	&81.3	&71.3	&47.5	&69.4 & \multirow{-2}{*}{\textcolor{red}{(+3.6)}}                                          \\
\midrule \bottomrule
\end{tabular}
}%As described in ~Sec.\ref{sec:episodic_eval}, we divide all methods  }
\end{table*}
\section{Methodology}
\subsection{Online Computation of $\mathcal{C}(g)$}
Section~\ref{sec:real_time_tta} discussed the online evaluation protocol of TTA methods.
Here, we give more details on the calculation of $\mathcal{C}(g)$, the relative adaptation speed of $g$, during our online evaluation.
First, we set $R\left(g(x)\right)$ as the time recording function for $g$ to perform a forward pass for a single batch.
To ensure a reliable time calculation, we execute \texttt{torch.cuda.synchronize()} before starting the timer and before ending it.
This ensures all GPU operations are finished for the moment time is computed.
To alleviate hardware dependence, we also calculate $R(f_\theta(x))$ for \textit{each} evaluation step computing the relative adaptation complexity. 
It is worth mentioning that $\mathcal C(g)$ for SHOT, EATA, SAR, and COTTA are $[3, 3, 8, 103]$ on average, respectively.
% For each evaluation step, 

\section{Experiments}
\subsection{Episodic Evaluation of TTA}
\paragraph{SHOT, PL, and ETA}
For completeness, we report the results on 3 baselines: Pseudo Label~\cite{pl}, SHOT-IM~\cite{shot}, and ETA~\cite{eata} in Table~\ref{tab:app_episodic}.
We follow the same setup as in the main paper.
Our results are consistent with the findings of Section~\ref{sec:episodic_eval} and Table~\ref{tab:imagenetc_episodic}.
In particular, SHOT-IM improves its performance under the online evaluation, similar to SHOT.
Further, the performance of ETA and PL degrades under the online evaluation due to the additional computational burden.
Nevertheless, ETA is similar to EATA in providing the best tradeoff between additional computational requirements and performance improvements.

\paragraph{SAR with GN} 
We equip our results to include ResNet50 with Group Normalization (GN) layers, following~\cite{sar}.
We report the results in Table~\ref{tab:app_episodic}, where we observe that:
\textbf{(i)} Under a relatively large batch size~(64), ResNet50 with GN underperforms ResNet50 with Batch Normalization. 
In fact, the average error rate for SAR increases from 56.2\% to 65.8\%.
\textbf{(ii)} The online evaluation penalizes SAR in both architecture choices with a performance degradation of 3.6\% under the GN-based ResNet.
Finally, it is worth mentioning that SAR with GN layers attains a similar performance under a batch size of 1.

\begin{figure}[t]
    \centering
    \includegraphics[width=1.0\linewidth]{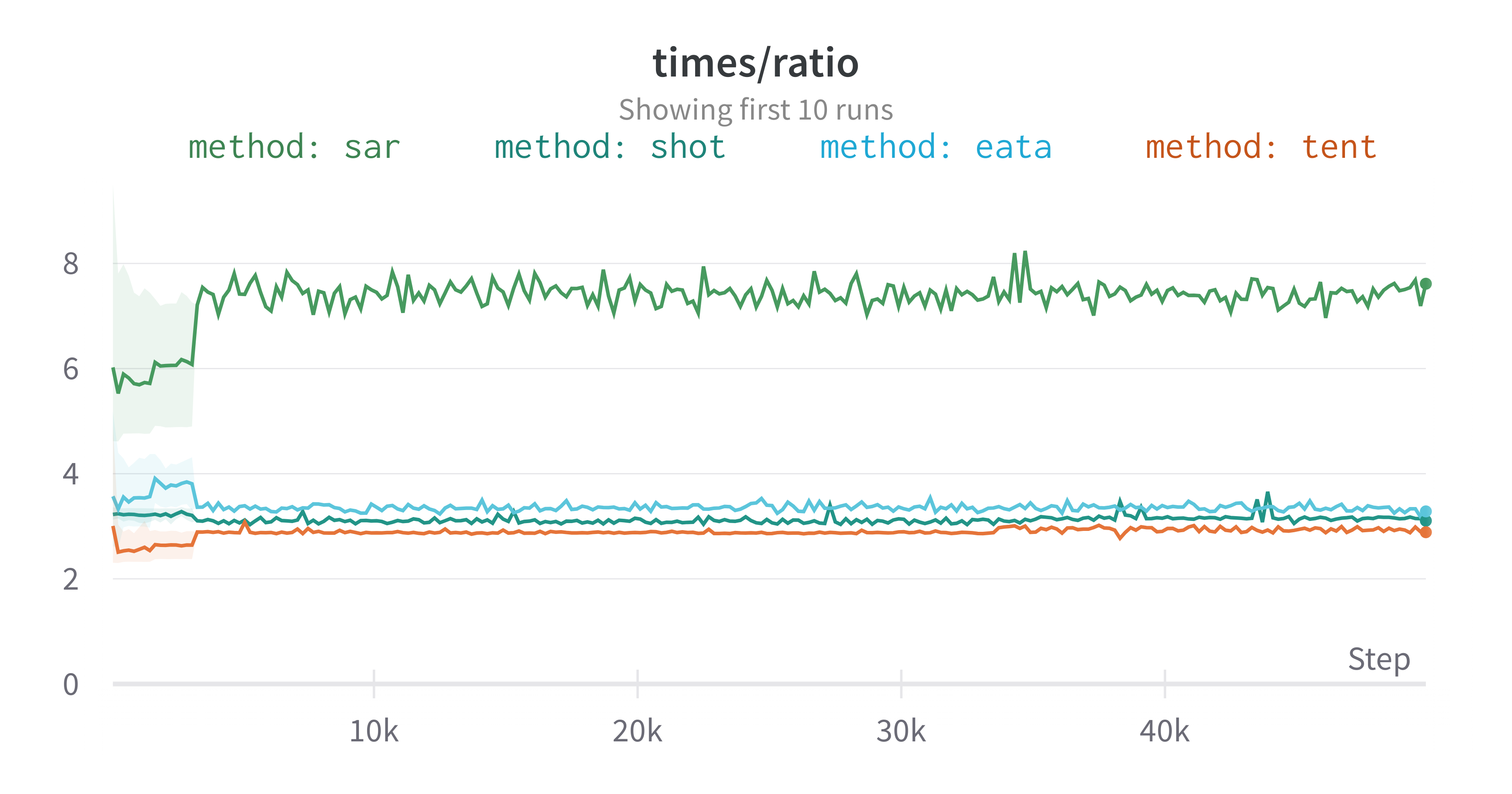}
    \caption{\textbf{$\mathcal C(g)$ computation across iterations.}
     We report our online calculations for the relative adaptation speed of~$g$, $\mathcal C(g)$, for SAR, SHOT, EATA, and TENT throughout a full evaluation episode.
    We observe that, overall, $\mathcal C(g)$ has a stable behavior throughout evaluation iterations.}
    \label{fig:cg_calculation}
\end{figure}

\begin{figure*}[t]
    \centering
    \includegraphics[width=1.0\textwidth]{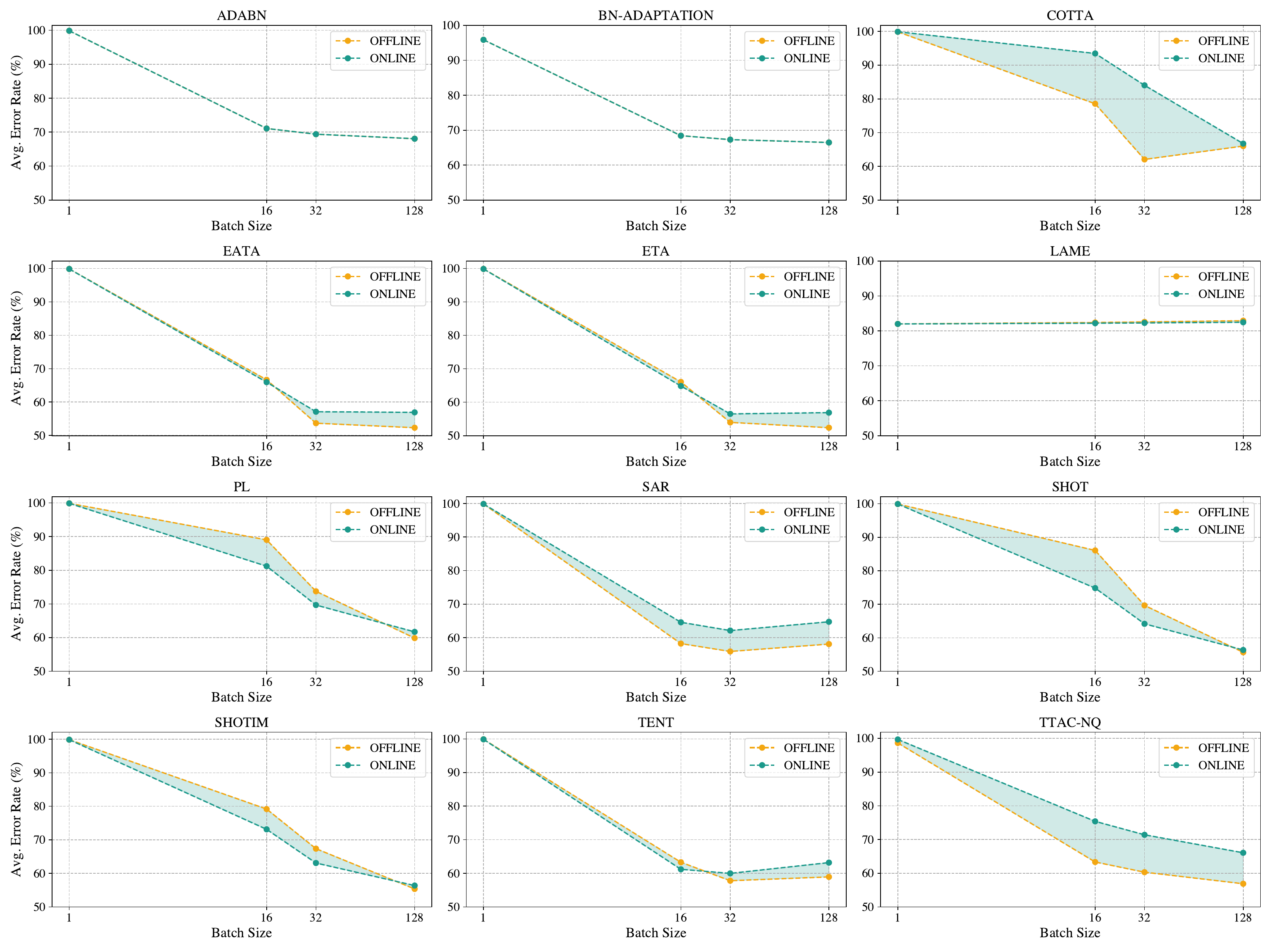}
    \caption{\textbf{Batch Size Analysis current \textit{vs.} realistic setups for every method.}
    We assess the performance variation of 12 different TTA methods under varying batch sizes. 
    We experiment with batch sizes in \{1, 16, 32, 128\}. 
    We do not include the baseline, MEMO, and DDA, since they are data-dependent approaches and are unaffected by batch size. 
    All TTA methods, except LAME, are severely affected by smaller batch sizes.
    Nonetheless, the realistic evaluation degrades the performance of all methods, except SHOT and SHOT-IM.
    }
    \label{fig:offline_online_batchsize_pairwise}
\end{figure*}

\paragraph{Ablating Batch Sizes} %Check scaling the learning rate
In the experiments section, we fixed the batch size to 64 following the recommendations of earlier works~\cite{tent, eata}.
Here, we investigate the effect of our proposed online evaluation under different choices of batch sizes.
To that end, we vary the batch size in \{1, 16, 32, 128\}, and report the results in Figure~\ref{fig:offline_online_batchsize_pairwise}.
We draw the following observations:

\textbf{(i) Online evaluation improves the performance of SHOT and SHOT-IM}.
This result is consistent with the earlier observations in Table~\ref{tab:imagenetc_episodic}. 
Note that PL shares a similar trend as well.

\textbf{(ii) The performance of TTA methods degrades when switching from offline to online evaluation, regardless of the batch size.} 
This result is highlighted in COTTA, ETA, EATA, SAR, TENT, and TTAC-NQ.

\textbf{(iii) Performance of TTA methods vastly varies when varying the batch size.} 
This result is consistent with earlier findings in the literature~\cite{dda, sar}, where most TTA methods fail with small batch sizes.

\begin{figure*}[t]
    \centering
    \includegraphics[width=1.0\textwidth]{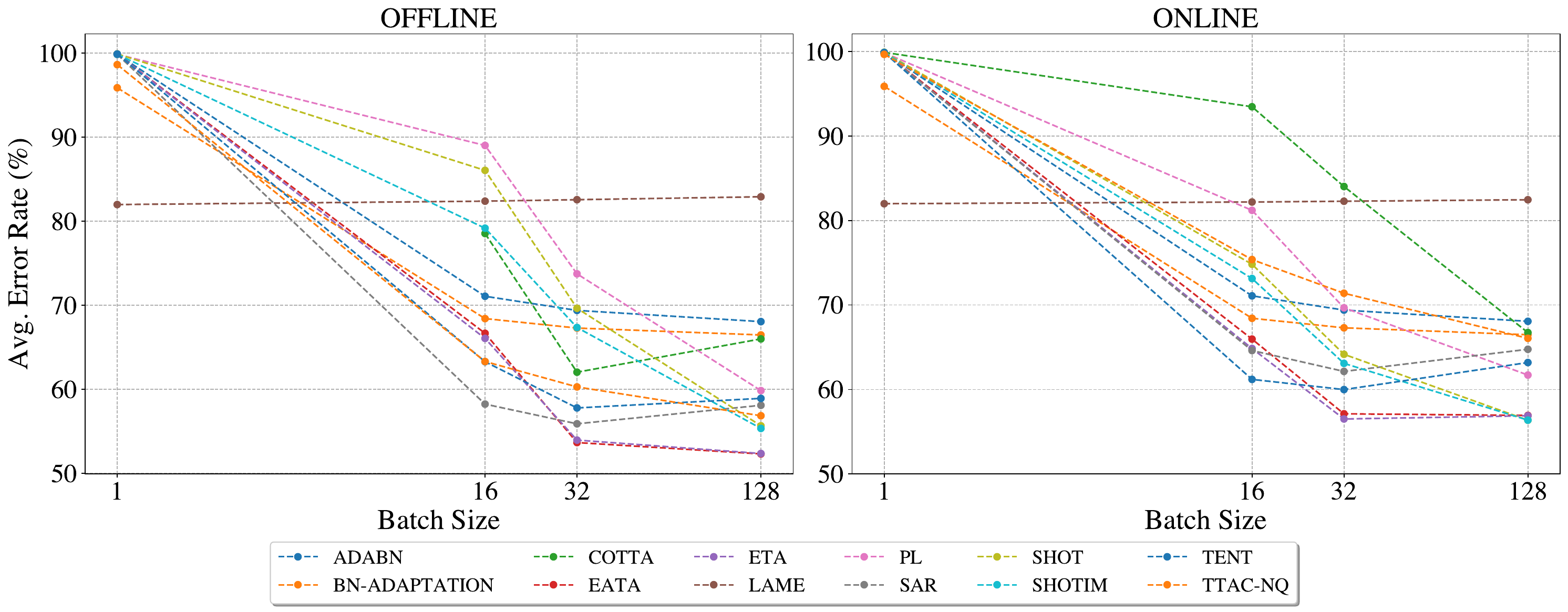}
    \caption{\textbf{Summary of batch size analysis: current \textit{vs.} realistic setups.} 
    Left: Current evaluation, \ie Section~\ref{sec:test time adaptation}. 
    Right: Realistic evaluation,\ie Section~\ref{sec:real_time_tta}. 
    While EATA achieves the lowest error rate under batch sizes $\geq 32$, SHOT becomes a very competitive baseline, outperforming EATA, at a batch size of 128.
    }
    \label{fig:offline_online_batchsize}
\end{figure*}

At last, and to ease comparison across methods, we summarize all the plots for all methods in Figure~\ref{fig:offline_online_batchsize}.

% \paragraph{C(g) Calculation over Iterations}

\paragraph{Consistency with 3 random seeds.}
For all of our experiments, we run each experiment with 3 random seeds. 
In most of our results, we found out that the standard deviation of performance across runs is very small. 
Our results in Figures~\ref{fig:continual_setup} and \ref{fig:stream_speed_standard} demonstrate this variation in the shaded area for 5 different TTA methods.

\subsection{Continual Evaluation of TTA}
We further explore another setup for the continual evaluation of TTA.
In particular, we follow~\cite{cotta} in concatenating all corruptions in ImageNet-C with 11 different orders.
We then report the average performance of each method across all runs and corruptions in Table~\ref{tab:app_continual}. 
We run each experiment with 3 random seeds, and report our results with standard deviations.
For the remaining implementation details, we follow our setup in main paper.
We observe that, similar to our conclusions in Section~\ref{sec:continual_eval}, online evaluation helps methods that do not perform sample rejection (\eg TENT).
Nonetheless, both ETA and EATA provide the best trade-off between performance and additional computational burden.

\begin{table*}[]
\caption{\label{tab:app_continual} \textbf{Continual Error Rate on ImageNet-C.}
We report the average continual error rate for 11 different corruption orders, with 3 different seeds, under both the offline and online setups with a corruption severity level of 5. 
\emph{Continual} refers to continually adapting after each corruption without resetting. 
This metric indicates the model's capability to learn from previous corruptions. 
The offline setup refers to the performance of the model in a continual learning scheme, whereas the online setup refers to the performance of the model in a continual learning scheme, under our more realistic online setup. 
We show that the more complex a method is, the fewer samples it adapts to, achieving better performance in a continual learning scheme.}
\centering
\begin{tabular}{l|cccccccc}
\toprule
{Avg. Error (\%)} &COTTA & ETA &  TENT & SAR & EATA & SHOT & TTAC-NQ \\\midrule
Offline&
$65.3\pm5.9$&
$56.4\pm2.3$&
$84.6\pm16.0$&
$59.8\pm3.0$&
$56.4\pm2.3$&
$88.4\pm11.4$&
$81.8\pm11.4$\\
\midrule Online&
$69.3\pm2.8$&
$57.7\pm2.0$&
$65.6\pm5.0$&
$60.4\pm1.8$&
$57.7\pm1.9$&
$78.2\pm7.7$&
$65.1\pm3.8$\\\bottomrule
\end{tabular}
\end{table*}
\begin{figure}[b]
    \centering
    \includegraphics[width=1.0\columnwidth]{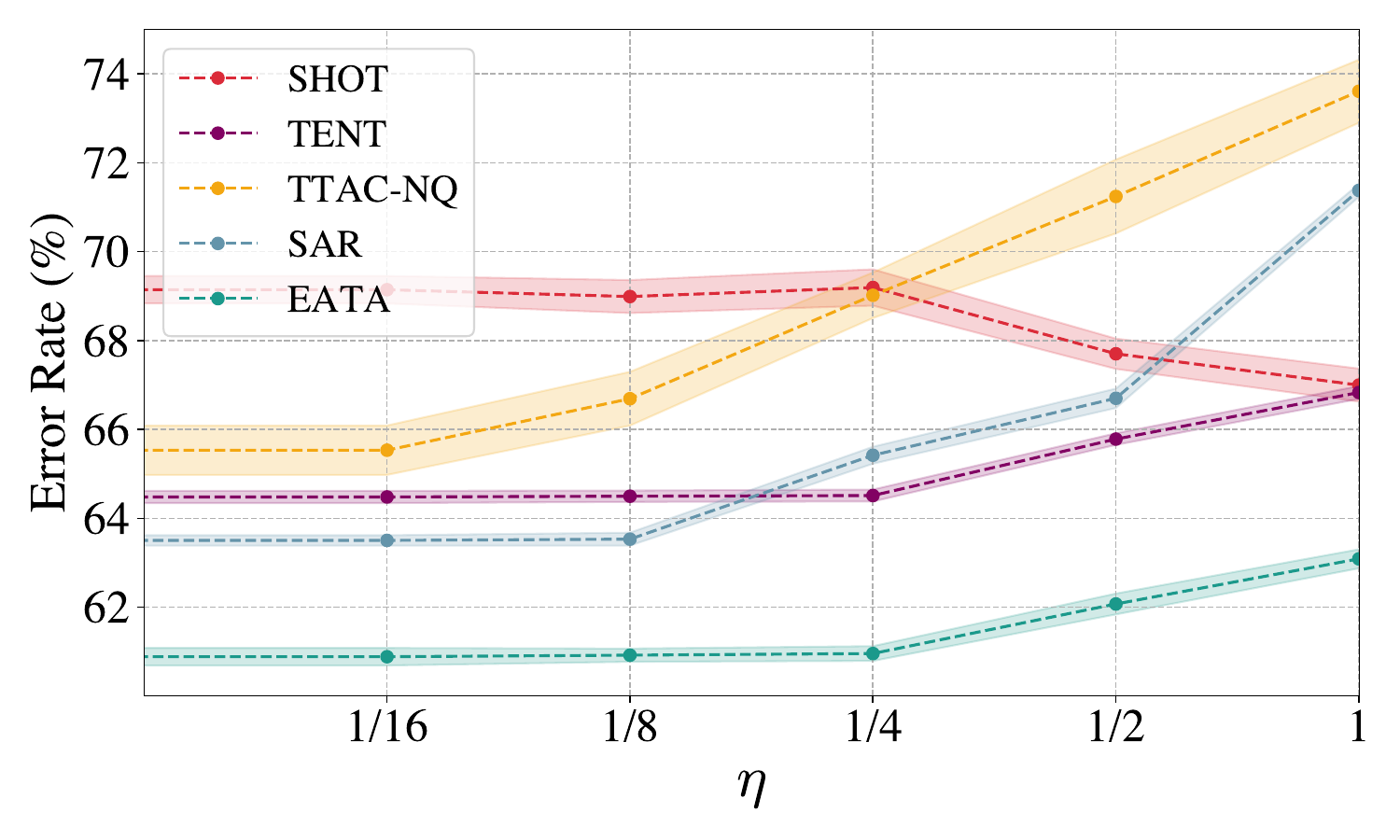}\vspace{-0.2cm}
    \caption{\textbf{Average Error Rate on ImageNet-3DCC Under Slower Stream Speeds.} 
    We report the average error rate for several TTA methods on ImageNet-3DCC under slower stream speeds. In our proposed online model evaluation, the stream speed $r$ is normalized by the time needed for a forward pass using the base model. 
    % However, this might not be true, depending on the application. 
    % The stream speed can actually be slower. 
    We evaluate different TTA methods under a stream with speed $\eta r$ with $\eta \in (0, 1]$. 
    % We define a ratio $\eta$ that controls the speed stream. 
    An $\eta=\nicefrac{1}{16}$ means the stream is $16$ times slower than the forward pass of the base model. 
    % If the stream is faster, \ie $\eta>1$, we argue that the reasonable course of action is to change the base model to a faster one. 
    We report the standard deviation across 3 random seeds.
    Different TTA methods degrade differently when varying $\eta$.
    % \juan{the $\eta$ in the xlabel looks too bold}
    }\vspace{-0.2cm}
    \label{fig:stream_speed_3dcc_standard}
\end{figure}

% \shyma{numbers will be edited very slightly we are waiting on 7 more jobs}
\subsection{Stream Speed Analysis}
For completeness, we extend our stream speed analysis in Section~\ref{sec:stream_speed} to cover the ImageNet-3DCC dataset. 
We preserve our experimental setup by varying the stream speed according to $\eta r$, with $\eta \in \{ \nicefrac{1}{16}, \nicefrac{1}{8}, \nicefrac{1}{4}, \nicefrac{1}{2}, 1$.
Figure~\ref{fig:stream_speed_3dcc_standard} summarizes our results for SHOT, TENT, TTAC-NQ, EATA, and SAR.
We observe similar trends to the ones in Figure~\ref{fig:stream_speed_standard}, where the performance of different TTA methods varies widely under different stream speeds.
The large relative adaptation speed of TTAC-NQ degrades its performance under even slow streams (\eg $\eta=\nicefrac{1}{8}$), while SHOT reduces its error rate under faster streams.
Furthermore, EATA is consistently outperforming all other considered approaches under different stream speeds.

\subsection{Evaluation on Other Benchmarks}

% \usepackage{multirow}
\begin{table*}[]
\caption{\textbf{Episodic Error Rate on ImageNet-3DCommonCorruptions.} 
We report the error rate of different TTA methods on ImageNet-3DCC \cite{3dcc} benchmark under both the realistic and offline setups. 
A lower error rate indicates a better TTA method. 
The highlighted numbers indicate a better performance per method across setups. 
Episodic means the model will adapt to one corruption at a time. 
The model is reset back to the base model when moving to the next corruption. 
The offline setup corresponds to reproducing the reported performance of every method. 
The first sub-table corresponds to methods that incur none or few additional computations, \ie $\mathcal{C}(g) = 1$. 
We show that methods generally perform worse in the more realistic setup. 
The more computationally complex the TTA method is, the fewer data it will adapt to, and the worse its performance.}
\label{tab:imagenet3dcc_episodic}
\resizebox{\textwidth}{!}{
% \usepackage{multirow}
\begin{tabular}{l|c|cc|ccc|c|c|ccc|cc|c|c}
\toprule
 \multicolumn{1}{c|}{} & \multicolumn{1}{c|}{} & \multicolumn{2}{c|}{Depth of field} & \multicolumn{3}{c|}{Noise} & \multicolumn{1}{c|}{Lighting} & \multicolumn{1}{c|}{Weather} & \multicolumn{3}{c|}{Video} & \multicolumn{2}{c|}{Camera motion} & \multicolumn{1}{c|}{} & \multicolumn{1}{c}{}\\
\multirow{-2}{*}{Method} & \multirow{-2}{*}{Realistic} &
\multicolumn{1}{c}{Near focus} & \multicolumn{1}{c|}{Far focus} & \multicolumn{1}{c}{Color quant.} & \multicolumn{1}{c}{ISO noise} & \multicolumn{1}{c|}{Low light} & \multicolumn{1}{c|}{Flash}    & \multicolumn{1}{c|}{Fog 3D}  & \multicolumn{1}{c}{Bit error} & \multicolumn{1}{c}{H.265 ABR} & \multicolumn{1}{c|}{H.265 CRF} & \multicolumn{1}{c}{XY-mot. blur} & \multicolumn{1}{c|}{Z-mot. blur}  & \multirow{-2}{*}{Avg.} & \multirow{-2}{*}{$\Delta$} \\
\midrule
                                % & \cmark                                  & 46.9                           & 55.6                          & 82.5                             & 94.0                          & 71.7                          & 78.7                         & 75.3                        & 88.6                          & 70.6                          & 65.4                          & 82.0                             & 75.3                            & 73.9                                                                  \\
\multirow{1}{*}{Source}         & \cmark                                     & 46.9                           & 55.6                          & 82.5                             & 94.0                          & 71.7                          & 78.7                         & 75.3                        & 88.6                          & 70.6                          & 65.4                          & 82.0                             & 75.3                            & 73.9   & -                                                               \\
\multirow{1}{*}{AdaBN} & \cmark                                  & 45.2                           & 55.0                          & 71.8                             & 76.8                          & 64.1                          & 80.8                         & 75.0                        & 91.8                          & 80.9                          & 76.7                          & 79.1                             & 67.5                            & 72.1 & -                                                                  \\     
% & Online                                     & 45.2                           & 55.0                          & 71.8                             & 76.8                          & 64.1                          & 80.8                         & 75.0                        & 91.8                          & 80.9                          & 76.7                          & 79.1                             & 67.5                            & 72.1                                                                  \\
\multirow{1}{*}{LAME}                                    & \cmark                                  & 45.3                           & 55.0                          & 71.9                             & 76.9                          & 64.1                          & 80.8                         & 75.1                        & 91.8                          & 80.9                          & 76.8                          & 79.2                             & 67.6                            & 72.1 & -                                                                 \\
% & Online                                     & 45.2                           & 55.0                          & 71.8                             & 76.9                          & 64.1                          & 80.8                         & 75.0                        & 91.8                          & 80.9                          & 76.8                          & 79.2                             & 67.5                            & 72.1                                                                  \\
\multirow{1}{*}{BN}                                & \cmark                                  & 43.9                           & 54.3                          & 72.3                             & 76.6                          & 60.9                          & 80.1                         & 72.4                        & 90.9                          & 78.7                          & 73.8                          & 76.9                             & 65.6                            & 70.5  & -                                                                 \\ 
% & Online                                     & 43.9                           & 54.3                          & 72.3                             & 76.6                          & 60.9                          & 80.1                         & 72.4                        & 90.9                          & 78.7                          & 73.8                          & 76.9                             & 65.6                            & 70.5                                                                  \\
\midrule
\midrule
\multirow{2}{*}{PL}                                     & \xmark                                  & \cellcolor[HTML]{e9f5f9}39.8                           & \cellcolor[HTML]{e9f5f9}49.8                          & \cellcolor[HTML]{e9f5f9}65.5                             & 72.6                          & \cellcolor[HTML]{e9f5f9}48.9                          & 79.0                         & \cellcolor[HTML]{e9f5f9}66.1                        & 97.5                          & 92.1                          & 86.2                          & 88.7                             & \cellcolor[HTML]{e9f5f9}57.6                            & 70.3 & \multirow{2}{*}{\textcolor{green}{(-1.6)}}                                                                  \\       & \cmark                                     & 41.0                           & 51.3                          & 66.5                             & \cellcolor[HTML]{e9f5f9}71.5                          & 52.8                          & \cellcolor[HTML]{e9f5f9}77.4                         & 68.1                        & \cellcolor[HTML]{e9f5f9}95.6                          & \cellcolor[HTML]{e9f5f9}86.0                          & \cellcolor[HTML]{e9f5f9}78.7                          & \cellcolor[HTML]{e9f5f9}77.0                             & 59.2                            & \cellcolor[HTML]{e9f5f9}68.7 &                                                                  \\
\midrule
\multirow{2}{*}{SHOT}                                    & \xmark                                  & 43.0                           & 53.6                          & 67.1                             & 64.2                          & 51.9                          & 81.1                         & 73.2                        & 97.2                          & 83.5                          & 77.8                          & 77.3                             & 60.1                            & 69.2 & \multirow{2}{*}{\textcolor{green}{(-2.2)}}                                                                  \\      & \cmark                                     & \cellcolor[HTML]{e9f5f9}41.7                           & \cellcolor[HTML]{e9f5f9}51.4                          & \cellcolor[HTML]{e9f5f9}64.4                             & \cellcolor[HTML]{e9f5f9}63.8                          & \cellcolor[HTML]{e9f5f9}51.6                          & \cellcolor[HTML]{e9f5f9}77.5                         & \cellcolor[HTML]{e9f5f9}71.6                        & \cellcolor[HTML]{e9f5f9}95.1                          & \cellcolor[HTML]{e9f5f9}79.9                          & \cellcolor[HTML]{e9f5f9}74.6                          & \cellcolor[HTML]{e9f5f9}73.7                             & \cellcolor[HTML]{e9f5f9}58.5                            & \cellcolor[HTML]{e9f5f9}67.0                                                                  \\
\midrule
\multirow{2}{*}{SHOT-IM } & \xmark                                  & 42.2                           & 52.7                          & 66.6                             & 63.7                          & \cellcolor[HTML]{e9f5f9}51.0                          & 81.0                         & 72.1                        & 97.0                          & 83.3                          & 77.6                          & 75.6                             & 59.2                            & 68.5 & \multirow{2}{*}{\textcolor{green}{(-1.9)}}                                                                  \\   & \cmark                                     & \cellcolor[HTML]{e9f5f9}\cellcolor[HTML]{e9f5f9}\cellcolor[HTML]{e9f5f9}\cellcolor[HTML]{e9f5f9}41.2                           & \cellcolor[HTML]{e9f5f9}\cellcolor[HTML]{e9f5f9}\cellcolor[HTML]{e9f5f9}51.2                          & \cellcolor[HTML]{e9f5f9}\cellcolor[HTML]{e9f5f9}64.4                             & \cellcolor[HTML]{e9f5f9}63.3                          & 51.3                          & \cellcolor[HTML]{e9f5f9}77.5                         & \cellcolor[HTML]{e9f5f9}70.9                        & \cellcolor[HTML]{e9f5f9}94.9                          & \cellcolor[HTML]{e9f5f9}79.4                          & \cellcolor[HTML]{e9f5f9}74.1                          & \cellcolor[HTML]{e9f5f9}72.3                             & \cellcolor[HTML]{e9f5f9}58.3                            & \cellcolor[HTML]{e9f5f9}66.6                                                                 \\
\midrule
\multirow{2}{*}{TENT } & \xmark                                  & \cellcolor[HTML]{e9f5f9}39.9                           & \cellcolor[HTML]{e9f5f9}49.6                          & \cellcolor[HTML]{e9f5f9}62.4                             & \cellcolor[HTML]{e9f5f9}62.2                          & \cellcolor[HTML]{e9f5f9}50.7                          & \cellcolor[HTML]{e9f5f9}75.6                         & \cellcolor[HTML]{e9f5f9}68.5                        & 91.6                          & \cellcolor[HTML]{e9f5f9}75.7                          & \cellcolor[HTML]{e9f5f9}70.2                          & \cellcolor[HTML]{e9f5f9}70.4                             & \cellcolor[HTML]{e9f5f9}57.0                            & \cellcolor[HTML]{e9f5f9}64.5   & \multirow{2}{*}{\textcolor{red}{(+2.3)}}                                                               \\       & \cmark                                     & 41.7                           & 51.4                          & 65.5                             & 67.2                          & 54.7                          & 77.4                         & 70.1                        & \cellcolor[HTML]{e9f5f9}90.7                          & 76.8                          & 71.9                          & 74.0                             & 60.8                            & 66.8                                                                  \\
\midrule
\multirow{2}{*}{SAR }                                     & \xmark                                  & \cellcolor[HTML]{e9f5f9}40.3                           & \cellcolor[HTML]{e9f5f9}50.0                          & \cellcolor[HTML]{e9f5f9}62.0                             & \cellcolor[HTML]{e9f5f9}61.2                          & \cellcolor[HTML]{e9f5f9}50.6                          & \cellcolor[HTML]{e9f5f9}73.8                         & \cellcolor[HTML]{e9f5f9}65.8                        & \cellcolor[HTML]{e9f5f9}90.1                          & \cellcolor[HTML]{e9f5f9}73.9                          & \cellcolor[HTML]{e9f5f9}68.8                          & \cellcolor[HTML]{e9f5f9}69.1                             & \cellcolor[HTML]{e9f5f9}56.8                            & \cellcolor[HTML]{e9f5f9}63.5 & \multirow{2}{*}{\textcolor{red}{(+6.9)}}                                                                 \\      & \cmark                                     & 44.9                           & 54.7                          & 71.1                             & 75.4                          & 62.6                          & 80.3                         & 73.8                        & 91.7                          & 80.5                          & 76.1                          & 78.6                             & 66.9                            & 71.4        
\\
\midrule
\multirow{2}{*}{ETA}  & \xmark                                  & \cellcolor[HTML]{e9f5f9}38.7                           & \cellcolor[HTML]{e9f5f9}47.9                          & \cellcolor[HTML]{e9f5f9}59.1                             & \cellcolor[HTML]{e9f5f9}56.7                          & \cellcolor[HTML]{e9f5f9}46.8                          & \cellcolor[HTML]{e9f5f9}71.0                         & \cellcolor[HTML]{e9f5f9}62.1                        & 90.6                          & \cellcolor[HTML]{e9f5f9}72.8                          & \cellcolor[HTML]{e9f5f9}67.3                          & \cellcolor[HTML]{e9f5f9}64.7                             & \cellcolor[HTML]{e9f5f9}52.9                            & \cellcolor[HTML]{e9f5f9}60.9 & \multirow{2}{*}{\textcolor{red}{(+2.3)}}                                                                  \\         & \cmark                                     & 39.7                           & 49.3                          & 61.6                             & 60.7                          & 50.0                          & 73.5                         & 65.2                        & \cellcolor[HTML]{e9f5f9}90.3                          & 74.4                          & 69.1                          & 68.8                             & 55.9                            & 63.2                                                                  \\
\midrule
\multirow{2}{*}{CoTTA} & \xmark                                  & \cellcolor[HTML]{e9f5f9}40.8                           & \cellcolor[HTML]{e9f5f9}50.9                          & \cellcolor[HTML]{e9f5f9}66.3                             & \cellcolor[HTML]{e9f5f9}68.3                          & \cellcolor[HTML]{e9f5f9}54.6                          & \cellcolor[HTML]{e9f5f9}77.2                         & \cellcolor[HTML]{e9f5f9}68.0                        & \cellcolor[HTML]{e9f5f9}90.2                          & \cellcolor[HTML]{e9f5f9}76.4                          & \cellcolor[HTML]{e9f5f9}71.1                          & \cellcolor[HTML]{e9f5f9}73.1                             & \cellcolor[HTML]{e9f5f9}60.4                            & \cellcolor[HTML]{e9f5f9}66.4 & \multirow{2}{*}{\textcolor{red}{(+9.2)}}                                                                  \\      & \cmark                                     & 55.4                           & 63.1                          & 74.1                             & 77.0                          & 64.7                          & 83.4                         & 78.1                        & 93.7                          & 84.0                          & 80.3                          & 81.7                             & 71.9                            & 75.6                                                                  \\
\midrule
\midrule
\multirow{2}{*}{TTAC-NQ} & \xmark                                  & \cellcolor[HTML]{e9f5f9}\cellcolor[HTML]{e9f5f9}40.7                           & \cellcolor[HTML]{e9f5f9}50.5                          & \cellcolor[HTML]{e9f5f9}61.0                             & \cellcolor[HTML]{e9f5f9}61.1                          & \cellcolor[HTML]{e9f5f9}51.5                          & \cellcolor[HTML]{e9f5f9}72.8                         & \cellcolor[HTML]{e9f5f9}66.6                        & \cellcolor[HTML]{e9f5f9}93.8                          & \cellcolor[HTML]{e9f5f9}81.1                          & \cellcolor[HTML]{e9f5f9}74.7                          & \cellcolor[HTML]{e9f5f9}75.7                             & \cellcolor[HTML]{e9f5f9}59.1                            & \cellcolor[HTML]{e9f5f9}65.7 & \multirow{2}{*}{\textcolor{red}{(+7.9)}}                                                                  \\     & \cmark                                     & 49.9                           & 57.0                          & 69.3                             & 72.3                          & 58.9                          & 79.8                         & 76.3                        & 95.8                          & 86.5                          & 83.0                          & 84.6                             & 69.8                            & 73.6                                                                  \\
\midrule
\multirow{2}{*}{EATA } & \xmark                                  & \cellcolor[HTML]{e9f5f9}38.6                           & \cellcolor[HTML]{e9f5f9}47.8                          & \cellcolor[HTML]{e9f5f9}59.2                             & \cellcolor[HTML]{e9f5f9}56.6                          & \cellcolor[HTML]{e9f5f9}46.9                          & \cellcolor[HTML]{e9f5f9}71.2                         & \cellcolor[HTML]{e9f5f9}62.2                        & 90.9                          & \cellcolor[HTML]{e9f5f9}72.5                          & \cellcolor[HTML]{e9f5f9}67.4                          & \cellcolor[HTML]{e9f5f9}64.6                             & \cellcolor[HTML]{e9f5f9}52.9                            & \cellcolor[HTML]{e9f5f9}60.9 & \multirow{2}{*}{\textcolor{red}{(+2.2)}}                                                                 \\       & \cmark                                     & 39.8                           & 49.3                          & 61.6                             & 60.5                          & 49.9                          & 73.5                         & 64.8                        & \cellcolor[HTML]{e9f5f9}90.6                          & 73.7                          & 69.1                          & 68.6                             & 55.7                            & 63.1                                                                  \\

\bottomrule
\bottomrule
\end{tabular}
}
\end{table*}

We report the error rates on all corruptions of ImageNet-3DCC~\cite{3dcc}, along with the overall average error rate, in Table~\ref{tab:imagenet3dcc_episodic}. 
The conclusions we draw for ImageNet-3DCC~\cite{3dcc} are very similar to the ones observed on ImageNet-C~\cite{imagenetc} (in Section~\ref{sec:episodic_eval}). 
We observe that efficient methods, with $\mathcal C(g) = 1$, such as LAME and BN, maintain performance. Furthermore, the performance of some TTA methods~\cite{tent,sar,eata,cotta} degrades in the online setup, while others that use pseudo labeling~\cite{pl,shot} actually improve. 
This degradation seems to be directly proportional to the amount of data a method misses according to its $\mathcal C(g)$.

% \begin{itemize}
%     \item  $\mathcal{C}\left(g(x_t)\right)$ of remaining methods line 375
%     \item SAR layer norm
%     \item Table for ETA, SHOT-IM, and PL line  & line 622
%     \item seed ablation discussion for all main results line 481
%     \item cont. on other methods line 683 (probably won't get results  by then)
%     \item other dataset stream speed experiments line 740
%     \item extended table 3 benchmarks on other datasets line 749
% \end{itemize}
% \subsection{Implementation Details}
% For all considered baselines, we followed the original implementation released by the corresponding paper with the suggested hyperparameters.
% \red{Shall we put links to those github repos? If no, we can remove this last subsection all together}

%%%%%%%%%%%%%%%%%%%%%%%%%%%%%%%%%%%%%%%%%%%%%%
%%%%%%%%%%%%%%%%%%%%%%%%%%%%%%%%%%%%%%%%%%%%%%
%%%%%%%%%%%%%%%%%%%%%%%%%%%%%%%%%%%%%%%%%%%%%%
%%%%%%%%%%%%%%%%%%%%%%%%%%%%%%%%%%%%%%%%%%%%%%
\section{Single Model Evaluation Scheme}

In Section \ref{sec:real_time_tta}, we assume $f_{\theta_t}$ can generate predictions whenever $g$ is occupied with adapting to a batch. 
This setup assumes the capacity to concurrently deploy two models. %  at the same time.
%While one model is busy with the adaptation, the second model can generate predictions. 
However, this assumption might be unfair to methods with $\mathcal{C}(g) = 1$, since it allows expensive methods to skip batches without large penalties. %  and still perform better than these light methods. 
We thus also study the case where only one model can be deployed.

%assumption heavily penalizes methods that adapt with a single forward pass and can be as fast as the stream $\mathcal C(g) = 1$. by doing for instance a simple adaptation of the batch norm statistic \cite{adabn}. 
% To be fair to fast-adapting models, 
%We propose to evaluate the TTA adaptation using a single model at deployment time. 
Studying this setup requires establishing a policy on how samples missed by the TTA method $g$ are treated. 
That is, when $g$ is busy adapting, all skipped samples still must be predicted without access to $f_{\theta_t}$. 
Depending on the application, this prediction could leverage prior knowledge about the problem \emph{e.g.} temporal correlation across samples, or the bias of the distribution. %  towards some classes. 
In our setup, we consider the most strict scenario in which, whenever $g$ is busy, a random classifier generates predictions for the incoming samples. 
This naive design choice results from our evaluation on ImageNet-based datasets, %  from the fact that all the datasets that we use in our experiments are derived from ImageNet \cite{deng2009imagenet} 
which contain images whose classes display no bias nor temporal correlation. %  uniformly distributed classes. 
% However, when evaluating on a video stream, using the previous prediction makes more sense as the environment does not change much temporally in most of the cases. 
% Similarly, when evaluating on other datasets with long-tailed distribution, one could think of leveraging a blind classifier based on the history of predicted samples as the one proposed in \cite{cloc}.
We conduct episodic evaluation, similar to Section~\ref{sec:episodic_eval}, on ImageNet-C dataset.
We average the error rates per corruption category (\emph{e.g.} averaging error rates for gaussian, shot, and impulse noises) and present the results of this study in Table \ref{tab:imagenetc_single_model}.
We draw the following observation.

\textbf{Single model evaluation strongly favors methods with $\mathcal{C}(g) = 1$.}  
We observe that all models that are slower than the stream are heavily penalized to the point that using the original pre-trained model becomes a better alternative.
However, methods that can be as fast as the stream, like AdaBN or BN, become the best alternative due to their speed.
This result encourages more research toward developing efficient TTA methods that have negligible additional computational overhead.

\begin{table}[t]
\caption{\label{tab:imagenetc_single_model} \textbf{Per Corruption Category Average Error Rate Using Single Model Evaluation on ImageNet-C.} We report the average error rate per corruption category of different TTA methods under single model realistic evaluation mode on ImageNet-C. Single model mode assumes the deployment of a single model $g$ instead of two under a constant speed stream $\mathcal{S}$. We assume the most extreme scenario, that is if a model $g$ is occupied adapting to a batch, the incoming batch is fed to a random classifier. We observe that the best TTA methods to use in this scenario are AdaBN \cite{adabn} and BN \cite{bnadaptation}, which simply adapt the BN statistics.}\vspace{-0.3cm}
\center
\footnotesize
{
\setlength{\tabcolsep}{4.7pt}
\resizebox{\columnwidth}{!}{
\begin{tabular}{lc|c|c|c|c|g}
\toprule
Method & Realistic & Noise & Blur & Weather  & Digital & Avg.  \\ 
\midrule 
Source & \cmark & 97.7 & 83.8 & 69.1  & 81.4 & 82.0 \\
% \midrule 
AdaBN & \cmark & 84.5 & 76.1 & 54.9 & 62.7 & 68.5 \\
BN  & \cmark & 84.1	& 73.1 & 54.2 & 59.9 & 66.7 \\
\midrule 
SHOT  & \cmark & 92.6 & 91.3 & 87.0 & 88.5  & 89.7 \\
TENT  & \cmark & 91.9 & 89.4 & 83.0 & 85.0  & 87.0 \\
SAR & \cmark & 95.6 & 94.0 & 90.1 & 91.3  & 92.6 \\
\midrule
EATA  & \cmark & 89.4 & 87.6 & 82.0 & 83.2  & 85.3 \\
TTAC-NQ & \cmark & 96.6 & 96.9 & 96.3 & 96.4  & 96.5 \\
                         \bottomrule
\end{tabular}}
}\vspace{-0.2cm}
\end{table}

\section{Results on ResNet18}
In our experiments in the main paper, we focused on the standard ResNet18-architecture, following the common practice in the literature. 
Here, and for completeness, we extend our results to cover the smaller and more efficient ResNet18 architecture.
Teble~\ref{tab:resnet18} summarizes the episodic evaluation of 6 TTA methods on ImageNet-C dataset.
Similar to our conclusions in the episodic evaluation section in the main paper, more expensive adaptation methods degrade more under our realistic evaluation scheme.
\begin{table}[t]
\caption{Evaluating different TTA methods with ResNet-18 architecture on ImageNet-C. We report the average error rate across all different types of corruptions (lower is better). TTA methods generally perform worse in the more realistic setup. The more computationally complex the TTA method is, the less data it will adapt to, and the worse is its performance.}
    \centering
    \small
    \resizebox{\columnwidth}{!}{
    \begin{tabular}{c|cccccc}
    \toprule
        Method & Basic & BN & SHOT & Tent & EATA & SAR \\
        \midrule
        Current & 85.4 & 70.1 & 64.4 & 64.9 & 59.7 & 63.8 \\
        Realistic & 85.4 & 70.1 & 64.5 & 68.3 & 63.2 & 69.5 \\
        \midrule
        Diff & - & - & 0.1 & 3.4 & 3.5 & 5.7 \\
        \midrule
        \bottomrule
    \end{tabular}}
    \label{tab:resnet18}
\end{table}
\end{document}